%% file: main.tex
\documentclass[11pt, a4paper]{anu}

\input{macros_imports}

\usepackage[numbers]{natbib}
\usepackage[dvipsnames]{xcolor}
\usepackage{subcaption}
\usepackage{etoolbox}
\usepackage{afterpage}
\usepackage{adjustbox}

\makeatletter
\def\ps@firstpage{%
  \let\@oddhead\@empty
  \let\@evenhead\@empty
\def\@oddfoot{%
    \vbox to 0pt{\vss
      \color{gray}%
      \hrule width 0.2\textwidth \vskip 5pt
      \hbox{\footnotesize $^\dagger$ Project Lead. \emph{Note: Authors with technical contributions are listed alphabetically by their first name.}}
    }%
  }%
  \let\@evenfoot\@oddfoot
}
\makeatother

\renewcommand{\today}{}

\title{LPM 1.0: Video-based Character Performance Model}

\author{
\fontsize{9.5pt}{11.5pt}\selectfont
Ailing~Zeng$^\dagger$,
Casper~Yang,
Chauncey~Ge,
Eddie~Zhang,
Garvey~Xu,
Gavin~Lin,
Gilbert~Gu,
Jeremy~Pi,
Leo~Li,
Mingyi~Shi,
Shawn~Wang,
Sheng~Bi,
Steven~Tang,
Thorn~Hang,
Tobey~Guo,
Vincent~Li,
Xin~Tong$^\dagger$,
Yikang~Li,
Yuchen~Sun,
Yue~(R)~Zhao,
Yuhan~Lu,
Yuwei~Li,
Zane~Zhang,
Zeshi~Yang,
and Zi~Ye
\vskip 1pt
\textbf{\fontsize{10pt}{10pt}\selectfont Project Page: \href{https://large-performance-model.github.io}{large-performance-model.github.io}}
}

\let\cite\citep
\begin{abstract} 
Performance, the externalization of intent, emotion, and personality through visual, vocal, and temporal behavior, is what makes a character
alive. 
Learning such performance from video is a promising alternative to traditional 3D pipelines. However, existing video models struggle to jointly achieve high expressiveness, real-time inference, and long-horizon identity stability, a tension we call the \textbf{performance trilemma}.
Conversation is the most comprehensive performance scenario, as characters simultaneously speak, listen, react, and emote while maintaining identity over time. To address this, we present \emph{LPM~1.0} (Large Performance Model), focusing on single-person \emph{full-duplex} audio-visual conversational performance.
Concretely, we build a multimodal human-centric dataset through strict filtering, speaking--listening audio-video pairing, performance understanding, and identity-aware multi-reference extraction; train a 17B-parameter Diffusion Transformer (\emph{Base LPM}) for highly controllable, identity-consistent performance through multimodal conditioning; and distill it into a causal streaming generator (\emph{Online LPM}) for low-latency, infinite-length interaction. 
At inference, given a character image with identity-aware references, \emph{LPM~1.0} generates listening videos from user audio and speaking videos from synthesized audio, with text prompts for motion control, all at real-time speed with identity-stable, infinite-length generation. \emph{LPM~1.0} thus serves as a visual engine for conversational agents, live streaming characters, and game NPCs. To systematically evaluate this setting, we propose \emph{LPM-Bench}, the first benchmark for interactive character performance. \emph{LPM~1.0} achieves state-of-the-art results across all evaluated dimensions while
maintaining real-time inference.
\end{abstract}

\begin{document}
\maketitle

\thispagestyle{firstpage}

\begin{figure}[H]
    \centering
    \includegraphics[width=1.0\textwidth]{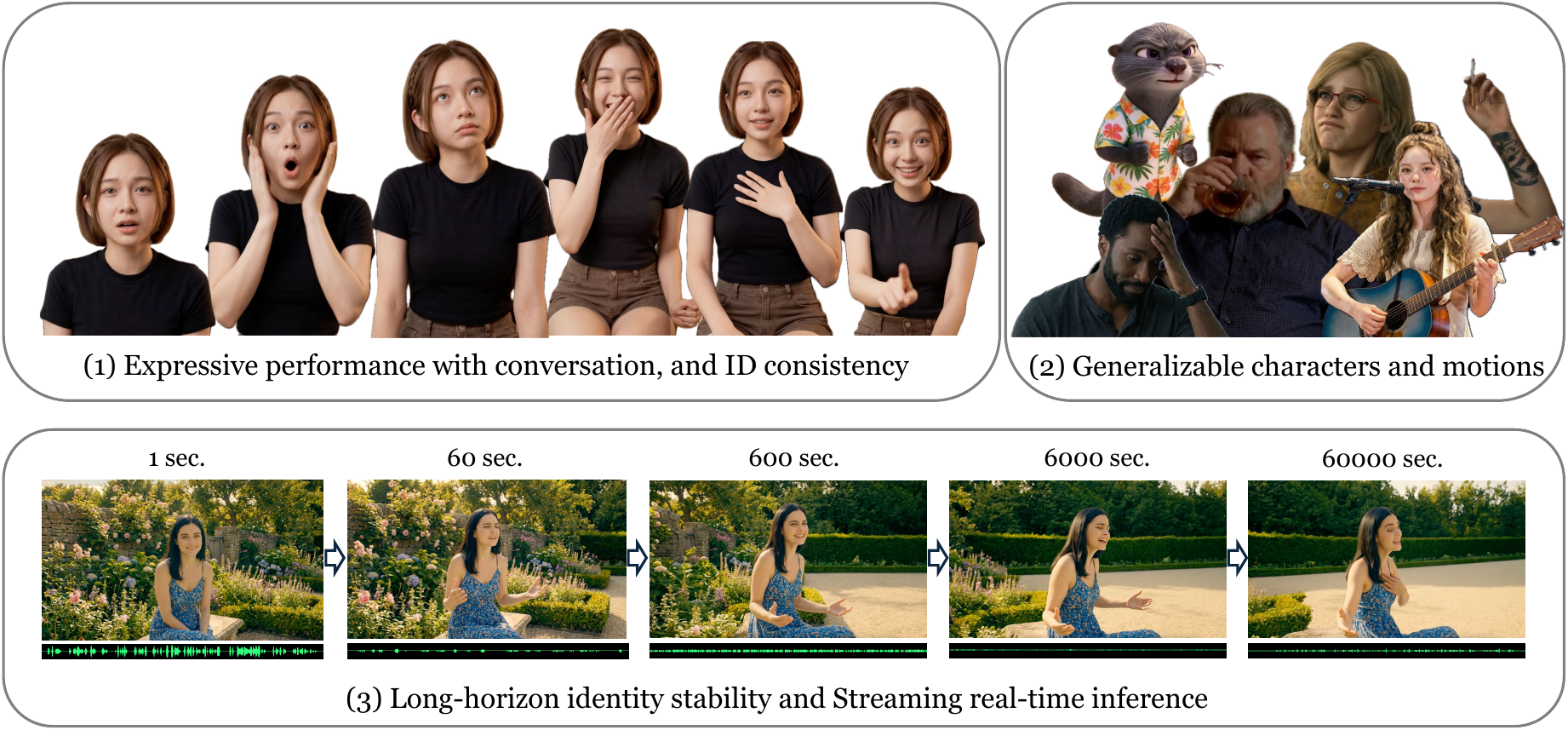}
    \caption{\textbf{LPM 1.0} generates identity-consistent conversational video with synchronized verbal and non-verbal behaviors---speaking, listening, micro-expressions, and natural motion---while maintaining visual fidelity across streaming and long-horizon video generation.}
    \label{fig:teaser}
\end{figure}

\newpage
\tableofcontents

\subimport{1_introduction}{1_introduction}

\subimport{4_dataset}{1_dataset_overview}
\subimport{4_dataset}{3_dataset_conversation}
\subimport{4_dataset}{4_dataset_caption}
\subimport{4_dataset}{5_dataset_ID}

\subimport{2_base_model}{1_base_architecture}

\subimport{2_base_model}{2_base_training}

\subimport{2_base_model}{3_base_inference}

\subimport{3_online_system}{1_online_architecture}

\subimport{3_online_system}{2_online_training}
\subimport{3_online_system}{3_online_inference}

\subimport{3.5_infra}{1_training_optimization}

\subimport{3.5_infra}{2_online_inference_optimization}

\subimport{5_eval_sections}{1_benchmark}

\subimport{5_eval_sections}{2_offline_compare}

\subimport{5_eval_sections}{3_online_compare}
\subimport{5_eval_sections}{4_ablation_study}

\input{7_discussion}

\subimport{8_safety_sections}{1_safety_security_responsibility}

\input{9_contributors}

\clearpage
\bibliographystyle{unsrt}
\bibliography{main}

\end{document}

%% file: macros_imports.tex
\usepackage{float}
\usepackage{makecell}        
\usepackage{fontawesome5}  
\usepackage{xspace}
\usepackage{multicol}
\usepackage{multirow}
\usepackage{longtable}
\usepackage{subcaption}
\usepackage{import}
\usepackage[usestackEOL]{stackengine}
\usepackage{framed}
\usepackage{graphicx}
\usepackage{svg}

\definecolor{headerbg}{RGB}{25, 25, 25}       
\definecolor{gridgray}{gray}{0.85}           
\definecolor{goodgreen}{RGB}{28, 142, 64}    
\definecolor{badred}{RGB}{237, 34, 39}       
\definecolor{cclnotreachedcolor}{RGB}{220, 53, 69} 

\definecolor{TargetGreen}{RGB}{46, 139, 87}

\definecolor{TargetGreen}{RGB}{46, 139, 87}

\newcolumntype{C}{>{\centering\arraybackslash}X}



%% file: 1_introduction/1_introduction.tex
\section{Introduction}
\label{sec:introduction}
\begin{flushright}
    \itshape ``Acting is reacting.'' --- \textsc{Sanford Meisner}
\end{flushright}
\vspace{0.5em}

Performance is fundamentally reactive: it arises through the continuous, embodied coordination of behavior in social interaction~\cite{goffman2023presentation}. It is conveyed not through words alone, but through rhythm, breath, hesitation, interruption, gaze, posture, gesture, and countless micro-expressions that reveal attention, intent, and emotion~\cite{knapp1972nonverbal}. What makes an interaction feel natural is the ability to react moment by moment: to listen before speaking, anticipate turn transitions, and soften, intensify, or withhold a response as the situation evolves~\cite{sacks1974simplest}. A \emph{performance model} should therefore do more than animate speech. Given a character specification and time-varying multimodal control signals, it must sustain an actor that \emph{speaks, listens, reacts, emotes, and moves} over time. In this work, we focus on bringing such performance to virtual characters in interactive conversation.

Today, production-quality character performance is still delivered primarily through conventional 3D pipelines involving modeling, rigging, animation, and rendering~\cite{parent2012computer}. While powerful due to their explicit modularity and controllability, these pipelines~\cite{zhu2023human} are inherently difficult to scale to the open-ended settings we target. Extending coverage across \emph{identities} (new characters), \emph{behaviors} (new styles of acting, emotion, and interaction context), and \emph{modalities} (faithful correspondences among audio, facial expression, body motion, and visual appearance) typically demands proportional increases in assets, capture data, and expert labor. As interactive applications proliferate—from human-like agents~\cite{park2023generative} to massive open-world conversational NPCs~\cite{nvidia2024ace}—the key challenge is no longer crafting a single high-end performance, but producing \emph{diverse} performances for \emph{many} characters under \emph{various} scenarios, quickly, reliably, and with minimal per-character authoring. Recent 3D-based efforts have begun exploring interactive conversational scenarios~\cite{chu2025unils,peng2026dyadit,cai2025towards}, yet they inherit the fundamental limitations of 3D pipelines.

This motivates a fundamentally different route: learning character performance directly from video in pixel space. Large-scale video data naturally contains broad variation in identities, viewpoints, conversational contexts, expressive emotions, and acting styles, suggesting that a single learned model could amortize authoring effort across characters and scenarios. At the same time, recent advances in diffusion-based video generation~\cite{wang2025wan,veo3,kling,Sora,seedance,LTXVideo} have substantially improved visual fidelity, while audio and text offer natural, low-barrier control signals for interactive use. Together, these developments make it plausible to synthesize character behavior from reference images, audio, and text \emph{without re-authoring character-specific rigs or capture pipelines}. 
However, existing video generation models face a surprisingly rigid constraint that we term the \textbf{performance trilemma}~\cite{lin2025omnihuman1,jiang2025omnihuman,team2025klingavatar,liveavatar,shen2025soulx,SekoTalk,Infiniteavatar,wang2026flowact}. In practice, existing models struggle to satisfy three critical desiderata simultaneously:
\begin{itemize}[noitemsep,topsep=0pt,leftmargin=*]
    \item \textbf{Expressive quality} --- The ability to act as a real human, displaying rich, conversational, and non-repetitive motion and micro-behavior, and communicative gaze.
    \item \textbf{Real-time inference} --- The capacity for causal, real-time generation suitable for live streaming.
    \item \textbf{Long-horizon stability} --- The preservation of identity, anatomical structure, personalized styles, and visual fidelity over indefinite durations.
\end{itemize}
Optimizing for speed and stability often yields robotic, repetitive motion; prioritizing expressiveness typically breaks real-time responsiveness; while maintaining identity and fidelity over long horizons remains a challenge for autoregressive drift. Consequently, existing systems are competent talkers, yet poor performers.

Conversation stresses all three desiderata simultaneously: expressiveness spans from active speaking to attentive listening; inference is real-time to support turn-taking and reactive behavior; and identity remains stable across indefinite interaction horizons. 
We attribute the conversational performance limitations to four missing ingredients in current paradigms: \textbf{(1) listening behavior is largely absent}. Most models are exclusively speech-driven~\cite{lin2025omnihuman1,jiang2025omnihuman,team2025klingavatar,SekoTalk,low2025talkingmachines,speakervid,liveavatar,shen2025soulx,Infiniteavatar}, neglecting the critical role of the listener---attentive gaze, personalized reactions, anticipation, and turn-taking cues. These elements constitute half of natural conversation and are essential for conveying understanding and presence; \textbf{(2) multimodal controllability remains poor}. True performance demands the joint modeling of time-varying text instructions
(to direct facial expressions, body motions, and environment interactions), speaking audio (to drive lip synchronization, emotion, and rhythmic motion), and listening audio (to trigger reactive and understanding behaviors)~\cite{agrawal2025seamless,cai2025towards,guo2025arig,siniukov2025ditailistener,xie2025x,zhou2022responsive};
\textbf{(3) character specification is underspecified}. Conditioning on a
single reference image forces generative models to infer unseen details,
such as teeth, expression wrinkles, profile geometry, and body appearance
under novel viewpoints~\cite{team2025klingavatar,jiang2025omnihuman,gao2025wan}. Although these inferences may appear plausible on a
frame-by-frame basis, they often fail to remain consistent over long
sequences. This under-specification makes single-image conditioning
insufficient for production scenarios that require stable identity and
motion continuity; \textbf{(4) a human-centric video foundation model is missing}. Existing models are typically built on small-scale or domain-specific datasets~\cite{agrawal2025seamless,speakervid,openhumanvid,zhou2022vicox,zhang2021HDTF_data,zhu2022celebv_data,geng2023affective_data} without the benefit of large-scale, high-quality human-centric video pretraining~\cite{liveavatar,cui2026avatarforcing,li2025joyavatar,shen2025soulx,wang2026flowact,xiao2025knot,tu2025stableavatar,xu2024vasa}. The absence of such a foundation leads to limited visual fidelity, weak generalization across diverse identities and scenarios, poor multimodal controllability, and inadequate conversational capability, preventing these systems from functioning as general-purpose
performance engines.

To address the performance trilemma, \emph{LPM~1.0} provides a full-stack framework for \emph{video-generative conversational performance}, spanning high-quality data construction, base model training with multimodal conditioning, and online distillation for real-time deployment. We argue that the performance trilemma is not merely an architectural challenge, but a system-level one. Addressing it requires coordinated design across data, control, and inference: the data must cover a broad range of speaking, listening, and socially reactive behaviors; multimodal conditioning must render this behavioral space steerable at inference time; and deployment must preserve these capabilities under causal, low-latency streaming constraints.
We instantiate our framework through three ingredients. (1) We build a large-scale multimodal dataset through three dedicated pipelines: rigorous quality filtering, conversational audio-video paired data processing, and identity-aware reference image construction. These pipelines produce training data that is not only visually clean but also explicitly suitable for full-duplex conversational generation, including both speaking and listening behaviors and rich identity specification; (2) Building on this dataset and a 14B pretrained image-to-video foundation model~\cite{wang2025wan}, we introduce 3B additional parameters in the form of interleaved speak/listen audio cross-attention blocks, yielding a 17B model that we fully train the resulting model on several tens of millions of clips. This formulates \textbf{Base LPM}, which jointly learns four core capabilities: speech-driven temporal dynamics, content-aware listening reactions, fine-grained text-conditioned control, and multi-reference identity preservation. Base LPM generates controllable, high-quality character performance video with minute-level temporal coherence; (3) We then distill Base LPM into \textbf{Online LPM}, a causal streaming generator for low-latency, infinite-length interaction. We develop a four-stage autoregressive distillation curriculum that progressively transfers the visual quality and behavioral richness of the base model into a real-time backbone-refiner architecture: the backbone maintains temporally coherent latent trajectories, while the refiner recovers high-fidelity details. This separation enables Online LPM to preserve identity, synchronization, and motion realism over unbounded horizons under practical latency constraints. \emph{LPM~1.0} could thus serve as a visual engine for conversational agents, live streaming characters, and game NPCs.

To evaluate this new setting, we introduce \textbf{LPM-Bench}, the first benchmark designed for interactive character performance with multi-modal inputs.
LPM-Bench contains 1{,}000 test cases across five scenarios (speaking, listening, conversation, diverse human motion, and character generalization), each providing a high-resolution initial frame, multi-reference images, structured text directives, and paired speaking/listening audio.
The benchmark is constructed for systematic diversity along three axes:
\textbf{(1)~Appearance}: input images span varied camera shots, ethnicities, ages, artistic styles, and identity references;
\textbf{(2)~Performance}: text prompts cover 22 expression bases, 78 emotions, over 5{,}000 motion descriptors, and gaze/pose configurations;
\textbf{(3)~Audio}: input audio covers diverse timbres, emotions, languages, and durations from 5 seconds to 1 hour, with 10\% dedicated to long-form generation to validate temporal consistency over extended horizons.

Extensive experiments on LPM-Bench show that both Base LPM and Online LPM are consistently preferred over state-of-the-art models in human evaluations.
In pairwise comparisons, Base LPM (720P) is preferred over Kling-Avatar-2~\cite{team2025klingavatar} and OmniHuman-1.5~\cite{jiang2025omnihuman} by 64.3\% and 42.5\% of raters, with the largest margins on identity consistency and motion dynamics.
Online LPM (480P) is preferred over LiveAvatar~\cite{liveavatar} and SoulX~\cite{shen2025soulx} by 82.5\% and 64.1\%. Meanwhile, in direct comparison at matched resolution (480P), human raters judge Base LPM and Online LPM as indistinguishable in 42--88\% of cases across all evaluation dimensions and scenarios, indicating that real-time causal generation need not sacrifice perceived realism.

Our contributions are as follows:
\begin{enumerate}[leftmargin=1.5em]
    \item We present \emph{LPM~1.0}, the first video-generative system for single-person full-duplex conversational performance, providing a full-stack framework that jointly addresses expressiveness, real-time inference, and long-horizon stability.
    \item We construct a large-scale curated multimodal data foundation through rigorous quality filtering, conversational speaking and listening audio-video pairing, and identity-aware multi-reference extraction, enabling the learning of expressive and reactive character behavior at scale.
    \item We develop \textbf{Base LPM}, a 17B bidirectional Diffusion Transformer trained on over 1.7 trillion multimodal tokens, which jointly models speech-driven motion, listening behavior, text-guided performance control, and identity-preserving multi-reference conditioning.
    \item We introduce a multi-stage autoregressive distillation framework that converts Base LPM into \textbf{Online LPM}, a causal backbone--refiner streaming generator for real-time, infinite-length synthesis.
    \item We build \textbf{LPM-Bench}, the first benchmark for interactive character performance with complete multimodal inputs, and show that \emph{LPM~1.0} achieves state-of-the-art results across conversational naturalness, controllability, emotional performance, and long-term consistency.
\end{enumerate}

%% file: 4_dataset/1_dataset_overview.tex
\section{Data}
\label{sec:dataset}

Training a performance model for interactive video generation requires data that goes far beyond conventional talking-head corpora. Such a dataset must satisfy three key requirements. \textbf{First}, it provides broad \emph{visual and motion coverage} of high-quality human-centric videos, spanning diverse shot scales, motions, expressions, emotional states, and appearances, so that the model can generalize across the wide range of human visual presentation encountered in real-world settings. \textbf{Second}, it must capture rich \emph{conversational performance}, including both speaking and listening behaviors, multi-turn interaction dynamics, and reactive social signals, while covering the social factors that shape conversation, such as persona, relationship type, personality, topic, affect, and situational context. This diversity is essential for learning behavior that is not only visually plausible but also socially grounded and contextually appropriate. \textbf{Third}, it must support \emph{paired multimodal conditions} derived from raw video, including frame-aligned audio separated into speaking and listening streams, detailed captions for fine-grained motion control, and multi-granularity identity reference images to enforce character identity consistency and expression-dynamics fidelity over long-horizon generation.

Thus, we construct a scalable human-centric multimodal dataset tailored for conversation-oriented generative modeling through a multi-stage processing pipeline over diverse long-form videos. In contrast to prior large-scale efforts such as OpenHumanVid~\cite{openhumanvid} and SpeakerVid-5M~\cite{speakervid}, which primarily emphasize speak-only video generation, our pipeline targets high-quality \emph{conversational performance in the wild}, where conversational semantics, detailed text, and identity-consistent conditioning must be addressed jointly. Through aggressive filtering and distribution-aware balancing, the overall retention rate is kept below 10\%, yielding training clips with the quality, diversity, and annotation density required for high-fidelity conversational character generation.

\subsection{Preparing Human-centric High-quality Videos}

\paragraph{Data Collection.}
We curate a large-scale video corpus designed to balance \emph{diversity, coverage, and quality}. The collected data span a wide range of performance scenarios, with substantial variation in shot scale, body motion, facial expression, emotional state, and appearance attributes. To better capture full-duplex conversational behavior, the corpus also includes dialogue-driven recordings with clearly identifiable speaker--listener interactions, covering varied personas, interpersonal dynamics, and emotionally rich situations. All data undergoes a systematic quality and content audit by a large team of human annotators, who verify visual fidelity, content relevance, and conversational validity before the data enters the processing pipeline.

\begin{figure}[t]
    \centering
    \includegraphics[width=1\textwidth]{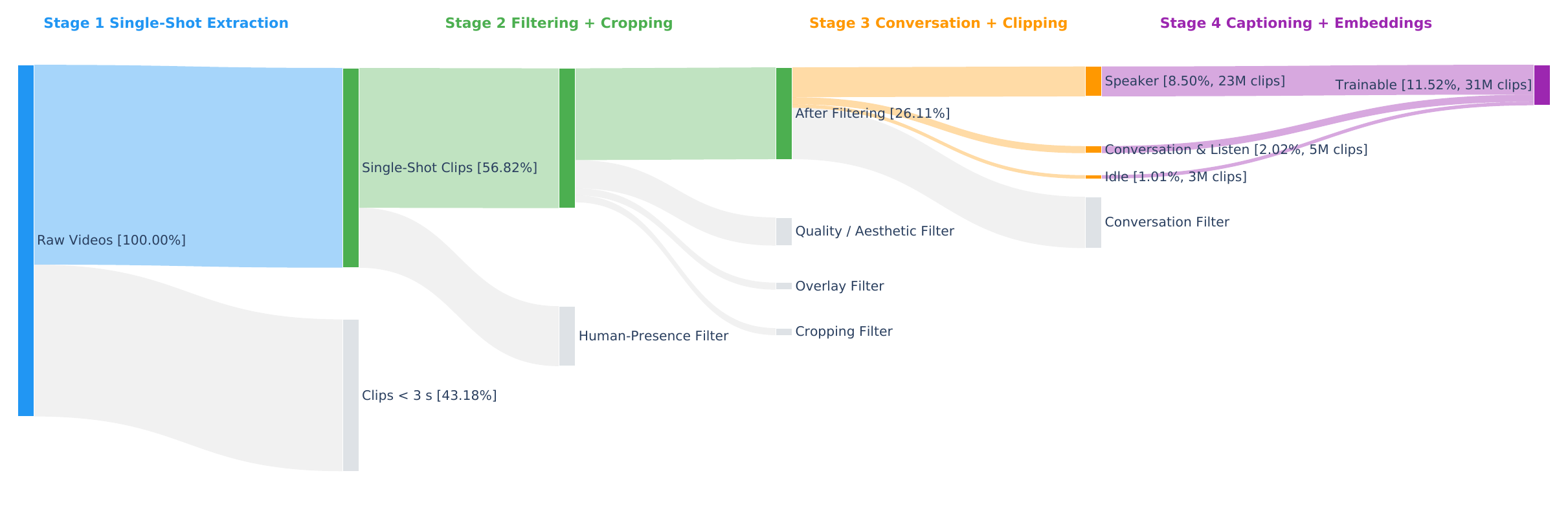}
    \caption{Data filtering and classification pipeline across four stages. Raw video is progressively filtered through single-shot extraction, quality filtering and cropping, conversation detection and clipping, and finally captioning with embedding generation to produce high-diversity, semantically rich, and emotionally expressive trainable clips.}
    \label{fig:video_filter_pipeline}
\end{figure}

\paragraph{Data Filtering and Classification.} The entire filter and classification processes are illustrated in Figure~\ref{fig:video_filter_pipeline}. Specifically, we convert the heterogeneous raw videos into standardized, high-quality single-shot clips through a multi-stage cleaning pipeline. Raw videos are first segmented via scene detection~\cite{transnetv2} into temporally coherent single shots, followed by a human detection model~\cite{wang2024yolov9} to remove clips without a visible person or excessive crowd density. 
To enhance audio-video quality, we apply a combined approach of manual inspection and model-based methods~\cite{finevq,qwen3vl,qwen3omni} to filter out problematic data across five major categories: (1) video editing and post-processing artifacts, such as transitions, jump cuts, frame drops, post-production overlays (logos, subtitles, watermarks, special effects) and beauty/skin-smoothing filters; (2) visual quality defects, including out-of-focus footage, global color shifts, and rapid screen flashing; (3) content authenticity issues, such as still images footage, AI-generated videos, or scenes with obviously composited/chroma-keyed backgrounds; (4) framing and composition issues, including incomplete human figures and dual-screen or picture-in-picture layouts; and (5) audio-video synchronization issues, covering desynchronized audio and videos where people's lips are moving without any audio. Through this pipeline, the proportion of quality-defective data identified during final manual quality inspection is kept below \textbf{1\%}.
Finally, retained clips are classified by the number of visible persons, producing structured metadata. The output is a corpus of high-quality, temporally coherent, person-count–annotated single-shot clips ready for semantic annotation. 
Building on this video corpus, we introduce three annotation modules that are particularly important for conversational character generation.

%% file: 4_dataset/3_dataset_conversation.tex
\subsection{Conversational Audio-Video Data Processing}
\label{sec:data_conv}

Conversational video depicts scenes in which characters engage in multi-turn visual-audio dialogue, alternating between speaking and listening.
To generate realistic conversational behavior, the model requires per-frame, per-character condition that explicitly labels each visible person as \emph{speaking}, \emph{listening}, or \emph{idle} at every moment---capturing not only speech production but also the non-verbal responses that characterize natural dialogue, such as nodding, smiling, changing gaze, or shifting attention.
%
Obtaining such labels at scale poses several challenges.
\textbf{(1) Conversation inference under limited observation.}
Determining whether a short clip depicts a conversation is non-trivial when only one side of the exchange is visible.
Such inference often requires multimodal reasoning: the speech content may suggest dialogue, while visual context provides supporting evidence (\emph{e.g.}, a partially visible interlocutor at the frame edge);
\textbf{(2) Defining listening across a broad behavioral spectrum.}
At one end, \emph{in-dialogue listening} manifests as attentive responses, such as nodding, smiling, or head-shaking, which are tightly coupled to a conversational partner's speech.
At the other end, \emph{general audio-reactive behavior} includes responses driven by non-speech audio, such as swaying to music or following spoken instructions.
The unifying criterion is whether the visible person's behavior exhibits meaningful correspondence to the concurrent audio input, which demands joint audio-video understanding rather than simple silence detection;
\textbf{(3) Fragility of na\"ive heuristics.}
Labeling frames as 'listening' solely based on background speech often leads to false positives, as off-screen narration or unrelated noise lacks visual evidence of engagement.
Specifically, the frame-level state classification rests on two foundational signals: \emph{speech detection} to identify when and whose speech is present (ideally with speaker-identity discrimination), and \emph{audio-visual synchrony} to determine whether a visible person's lip activity corresponds to the detected speech.
The classification logic follows from the joint state of these two signals.
Within detected speech segments: a person whose lip motion is synchronized with the audio is classified as \emph{speaking}; a person whose lips remain still is classified as \emph{listening}; and a person whose lip motion is present but desynchronized from the audio is flagged as inconsistent data and discarded.
Outside speech segments: a person with still lips is classified as \emph{idle}; a person with unexplained lip activity is similarly discarded.
Existing active speaker detection (ASD) methods~\cite{asd1,asd2,lrasd} and audio-visual alignment model (\emph{e.g.}, SyncNet~\cite{syncnet}) are designed to identify \emph{who is speaking} and do not model listening or idle states, verify audio-visual consistency for quality control, or support character-role assignment in multi-person conversational scenes.

\begin{figure}[tb]
    \centering
    \includegraphics[width=1.0\textwidth]{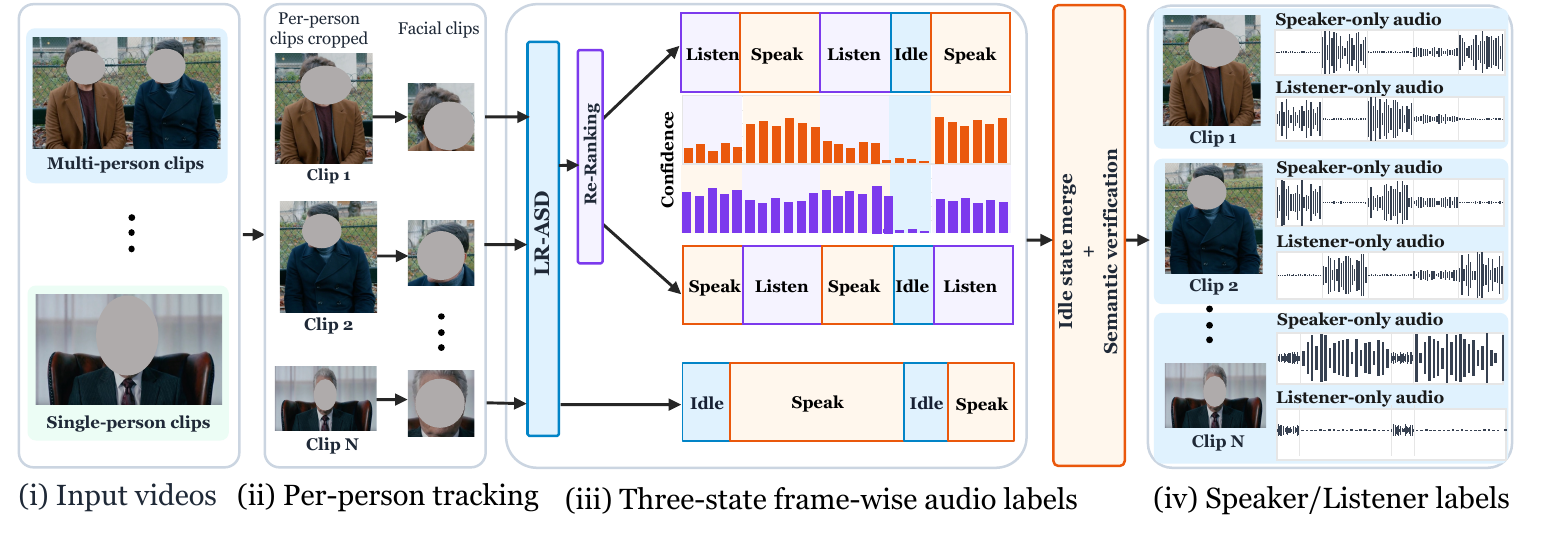}
      \caption{\textbf{Illustration of the conversational audio-video data processing pipeline.} \textbf{(1)} \emph{Tracking and cropping} converts multi-person clips into single-person clips;
      \textbf{(2)} \emph{Three-state labeling} applies fine-tuned LR-ASD to produce frame-wise
      \emph{speak/listen/idle} states;
      \textbf{(3)} \emph{Refinement and audio separation} verifies and filters labels, then outputs speaker/listener-only audio tracks for retained clips.
      }    \label{fig:conversation_fig}
\end{figure}

As shown in Figure~\ref{fig:conversation_fig}, we present the overall pipeline. We retain both single-person and multi-person clips; multi-person shots typically contain richer conversational dynamics (\emph{e.g.}, turn-taking and mutual attention) and therefore yield a larger fraction of useful training samples for conversation-oriented modeling.
Accordingly, we emphasize multi-person clips during conversation curation, while retaining single-person clips as complementary coverage.
Regardless of person count, each clip first undergoes person detection and tracking to obtain temporally consistent bounding-box trajectories.
For single-person clips, the full frame constitutes the sole person-centric track; for multi-person clips, the trajectories are used to crop individual \emph{person-centric} views, and low-fidelity tracks are removed by quality filtering.
We then perform frame-level audio state labeling on each person-centric track independently, producing per-person state predictions.
For multi-person clips, a character assignment and audio separation stage resolves speaker-listener roles and constructs character-conditioned audio variants; the clip is subsequently decomposed into per-person segments, each paired with the corresponding audio variant.
After decomposition, we further input these clips into an audio-video understanding model to classify the conversation scenarios.

\paragraph{Frame-level Audio State Labeling.}
\label{sec:data_process_labeling}

We finetune an ASD backbone (LR-ASD~\cite{lrasd}) to a three-state frame-level formulation via transfer learning. For the training set, we annotate an internal dataset of 20K single-shot clips (approximately 95 hours) with manual frame-level annotations, which substantially improves the separation of \emph{listening} versus \emph{idle} beyond standard ASD settings.
The adapted model jointly leverages speech detection and audio-visual synchrony to implement the classification logic described above: labeling each frame as speaking, listening, or idle while flagging audio-visually inconsistent frames for removal.
For both single-person and multi-person clips, we run the model on each person track to obtain per-person probability sequences over time.

\paragraph{Character-Audio Separation and Assignment.}
\label{sec:data_process_charassign}

We introduce a rule-driven, frame-level re-ranking module that aggregates per-person probabilities temporally and enforces consistency constraints.
Concretely, the re-ranking procedure: (1) compares candidate tracks frame by frame using speaking/listening confidence; (2) applies temporal smoothing and persistence checks to suppress transient spikes; (3) resolves character assignment by favoring temporally consistent speaker-listener patterns across tracks (\emph{e.g.}, stable speaker segments and coherent turn transitions).
Based on the resolved character assignment, we construct two character-conditioned audio variants for each multi-person clip: a \emph{speaker-only} track obtained by suppressing speech attributed to the listener, and a \emph{listener-only} track obtained by suppressing speech attributed to the speaker; \emph{idle} regions are retained in both tracks.
The clip is then decomposed into per-person segments, each paired with the corresponding audio variant, so that all subsequent processing operates uniformly on single-person inputs.

\paragraph{Semantic Verification via Audio-video Understanding Model.}
\label{sec:data_process_semverify}

While frame-level audio state labeling provides fine-grained behavioral annotations, it relies primarily on low-level audio-visual temporal cues and can produce systematic false positives for the listening state.
Two failure modes are particularly common: (1) \emph{false listening}, where the labeling model assigns a listening label to a segment that semantically corresponds to speaking (\emph{e.g.}, due to brief pauses or low-energy speech); and (2) \emph{not-in-dialogue listening}, where speech is present in the audio but the visible person does not exhibit meaningful behavioral correspondence to the concurrent speech (\emph{e.g.}, off-screen narration or background conversation).
To mitigate these errors, we introduce an additional semantic verification stage based on a fine-tuned audio-video understanding model~\cite{qwen3omni} that filters the true listening, conversation, and speaking clips.
Through training on 15K manually annotated clips with task-specific prompting, the model achieves an overall F1 that exceeds Gemini 2.5 Pro~\cite{comanici2025gemini} by \textbf{11.2\%} relatively (\textbf{+7.90} absolute).

\paragraph{Quantitative Results of the Conversation Pipeline.} This part quantitatively validates the two core operators introduced in Section~\ref{sec:data_conv}: frame-level audio state labeling and semantic verification via an audio-video understanding model. All metrics are computed on held-out clips with human annotations.
We report per-domain breakdowns across two representative data domains (\emph{Domain\,1} and \emph{Domain\,2}) whose distribution characteristics differ substantially, so as to reflect the diversity of real-world conversational data.

For frame-level audio state labeling, we benchmark the fine-tuned LR-ASD model on binary speak-versus-listen classification after re-ranking and idle merging.
Annotators assign per-frame states (\emph{speak}, \emph{listen}, and \emph{idle}) to single-shot, person-centric clips from both domains.
For multi-person sources, cross-track consistency is enforced: person-centric clips originating from the same shot are paired, and their labels are jointly refined so that \emph{speak}, \emph{listen}, and \emph{idle} are mutually exclusive at every time step.
Clips containing overlapping speech (\emph{e.g.}, simultaneous speakers) are excluded to avoid label ambiguity.
As shown in Table~\ref{tab:lrasd_results}, the model tests on the 2K manually annotated test clips (1K clips each from Domain 1 and Domain 2), achieving frame-level accuracy of \textbf{89.75\%} on Domain\,1 and \textbf{87.63\%} on Domain\,2.
On Domain\,1, speak and listen recalls are well balanced (91.62\% versus 87.99\%), whereas on Domain\,2, a notable asymmetry emerges: speak recall reaches 94.05\% but listen recall drops to 81.05\%.
This gap confirms the challenge described in Section~\ref{sec:data_conv}: in domains with higher acoustic variability, ambient noise and off-screen speech cause a fraction of true \emph{listening} frames to be misclassified as \emph{speaking}.
This is precisely the systematic error mode that the downstream semantic verification stage is designed to suppress.

\begin{table}[t]
\centering
\caption{Frame-level results for speak-versus-listen classification after re-ranking and idle merging on 2K manually annotated test clips (1K clips each from Domain 1 and Domain 2).
The left block shows the confusion matrix: entries are \textbf{frame counts} with column-normalized percentages in parentheses (equivalent to per-class recall on the diagonal).
The right block reports per-class precision and F1 score.
Summary metrics appear in the bottom row of each domain.
\textit{Pred.}: predicted label; \textit{GT}: ground truth; \textit{Acc.}: accuracy.}
\label{tab:lrasd_results}
\begin{tabular}{llrrcc}
\toprule
\multicolumn{2}{c}{} & \multicolumn{2}{c}{GT} & \multirow{2}{*}{Precision\ (\%)} & \multirow{2}{*}{F1 score} \\
\cmidrule(lr){3-4}
\multicolumn{2}{c}{} & Speak & Listen & & \\
\midrule
\multicolumn{6}{l}{\textbf{Domain\,1}} \\
\multirow{2}{*}{Pred.}
  & Speak  & 44{,}256\;(91.62\%) & 6{,}148\;(12.01\%)  & 87.80 & 89.67 \\
  & Listen & 4{,}049\;(8.38\%)   & 45{,}055\;(87.99\%) & 91.75 & 89.83 \\
\cmidrule(lr){2-6}
  & Total (GT) & 48{,}305 & 51{,}203
  & \multicolumn{2}{c}{Acc.\,=\textbf{89.75}\quad Macro-F1\,=\,\textbf{89.75}} \\
\midrule
\multicolumn{6}{l}{\textbf{Domain\,2}} \\
\multirow{2}{*}{Pred.}
  & Speak  & 123{,}112\;(94.05\%) & 24{,}236\;(18.95\%) & 83.55 & 88.49 \\
  & Listen & 7{,}789\;(5.95\%)    & 103{,}669\;(81.05\%) & 93.01 & 86.62 \\
\cmidrule(lr){2-6}
  & Total (GT) & 130{,}901 & 127{,}905
  & \multicolumn{2}{c}{Acc.\,=\,\textbf{87.63}\quad Macro-F1\,=\,\textbf{87.56}} \\
\bottomrule
\end{tabular}
\end{table}

For semantic verification via audio-video understanding model, we evaluate Gemini 2.5 Pro~\cite{comanici2025gemini} on manually annotated data as a closed-source baseline, then compare it with our fine-tuned Qwen3-Omni~\cite{qwen3omni} as described in Section~\ref{sec:data_process_semverify}.
Table~\ref{tab:gemini_qwen_metrics} reports per-class and overall metrics. 
As shown in Table~\ref{tab:gemini_qwen_metrics}, fine-tuning yields an overall F1 of \textbf{78.37}, an absolute gain of \textbf{+7.90} over the Gemini baseline (70.47). The largest per-class gains appear on \emph{silence} (+19.40) and \emph{speak} (+10.67), the two categories most susceptible to the failure modes identified in Section~\ref{sec:data_process_semverify}.
Examining the underlying confusion patterns reveals where these gains originate:
the \emph{silence}-to-\emph{speak} misclassification rate drops from 24.4\% to 9.0\%, directly suppressing false-speaking artifacts from silent segments;
the \emph{conversation}-to-\emph{listen\_dialogue} leakage decreases from 12.7\% to 4.8\%, improving the purity of conversation-labeled clips.
A residual weakness is that \emph{listen\_dialogue} clips are over-predicted as \emph{conversation} (25.4\% after fine-tuning versus 19.1\% for Gemini), suggesting that further calibration on this boundary case could yield additional gains.
\begin{table}[t]
\centering
\caption{F1 scores of the six-category semantic verification model. Gemini serves as the closed-source baseline; Qwen3-Omni is fine-tuned on domain-specific conversational data. Metrics cover the five primary classes (excluding \textit{unknown}). $\Delta$ denotes the absolute improvement of Qwen3-Omni over Gemini.
Abbreviations: \textit{L\_dlg}\,=\,Listen\_dialogue, \textit{L\_nondlg}\,=\,Listen\_nondialogue.}
\label{tab:gemini_qwen_metrics}
\begin{tabular}{lcccccc}
\toprule
Model & Conversation & L\_dlg & L\_nondlg & Silence & Speak & Overall \\
\midrule
Gemini 2.5 Pro (baseline)
  & \textbf{76.80} & 67.58 & \textbf{64.76} & 63.96 & 69.88 & 70.47 \\
Qwen3-Omni (fine-tuned)
  & 76.22 & \textbf{76.48} & 56.77 & \textbf{83.36} & \textbf{80.55} & \textbf{78.37} \\
\midrule
$\Delta$
  & $-$0.58 & +\textbf{8.90} & $-$7.99 & +\textbf{19.40} & +\textbf{10.67} & +\textbf{7.90} \\
\bottomrule
\end{tabular}
\end{table}

%% file: 4_dataset/4_dataset_caption.tex
\subsection{Audio-Video Understanding and Captioning}
\label{sec:data_caption}

To support both base model training and online model training, we adapt an audio-video understanding model to annotate two complementary forms of training data for precise human-centric video understanding: fixed-length clips and sentence-bounded variable-length segments. For each clip or segment, the model generates both dense natural language captions and structured tags. These tags capture diverse visual, audio, and conversational attributes, including motion, emotion, environment, camera, and other cinematographic properties. To improve annotation precision, we introduce a comprehensive taxonomy that spans multiple visual dimensions, encompassing both human performance and environmental context. In particular, for human performance, the taxonomy specifies fine-grained categories for emotions and facial expressions. This structured design improves multimodal representation quality while supporting consistent annotation and controllable generation.
Before captioning, variable-length segments are constructed by partitioning videos into ASR-based sentence units. Concretely, we split transcripts using punctuation cues to preserve semantic coherence and align each sentence unit with its corresponding temporal span in the video.

\paragraph{Analysis of Listener Videos.}

In naturally occurring conversational video, the vast majority of screen time is allocated to the active speaker, with a smaller portion capturing two-person dialogue shots; only approximately 10\% of all conversational segments are framed on the listener, depicting the non-speaking party's real-time reactions.
This inherent imbalance makes listener-centric data both scarce and valuable for training models that generate realistic listening behavior.
From the full pool of approximately 3.5M listener clips, we perform fine-grained annotation along multiple dimensions: the input audio content and its emotional tone, the listener's visible reaction (facial expression, eye gaze, body motion, and overall energy level), and contextual attributes inferred from the conversational setting such as the relationship between the interlocutors and the listener's personality.
The resulting distribution reveals a strong skew toward passive, emotionally subdued listening. Both emotion and facial expression distributions are heavily concentrated in neutral or cognitive categories, which together account for over 70\% of all labels, while expressive reactions such as \emph{anger}, \emph{fear}, and \emph{distress} each fall below 3\%. Motion intensity is low across roughly 90\% of the data, dominated by static postures and subtle movements concentrated in the head and eyes.
In short, the raw listener data overwhelmingly reflects calm, still, inward-focused attentiveness, with emotionally rich or physically dynamic reactions being rare.

To counteract this imbalance, we curate approximately 470K clips exhibiting clear emotional reactions or active engagement for supervised fine-tuning, and apply targeted rebalancing across four visual axes: emotion, expression, energy, and motion.
The core strategy is to down-sample dominant neutral categories and up-sample rare but training-critical behaviors, including high-energy responses, pronounced facial expressions, and diverse body movements such as nodding, gesturing, and gaze shifting.
The resulting set is enriched with two distinctive temporal patterns that are especially valuable for learning dynamic listening behavior:
\emph{emotion contrast}, where opposing affective states co-occur within a single clip (\emph{e.g.}, alternation between \emph{amusement} and \emph{embarrassment}, or between \emph{gratitude} and \emph{sadness}), and \emph{expression oscillation}, where facial expressions undergo three-phase transitions around a stable baseline (\emph{e.g.}, a brief shift from \emph{smiling} to \emph{surprise} before returning to \emph{smiling}).
Finally, a key design choice concerns listener captioning.
During inference, listener video generation must begin as soon as the speaker's audio stream arrives, leaving insufficient time for a language model to produce temporally aligned frame-level descriptions.
The captions for listener clips are therefore restricted to time-invariant attributes, such as appearance, personality, and relational context, rather than moment-by-moment behavioral narration.
This design encourages the model to learn the mapping from audio signal to visual response directly, rather than relying on text as an intermediary for temporal dynamics.

%% file: 4_dataset/5_dataset_ID.tex
\subsection{Identity-aware Reference Images Pipeline}
\label{sec:data_ID}

Conditioning on a single reference image forces the model to hallucinate unseen details. In professional artistic creation and character-driven storytelling, such inconsistencies are unacceptable: the same character's smile should always reveal the same teeth, the same outfit should appear consistent when viewed from the side or rear, and identity-specific facial details should persist across diverse expressions throughout an extended conversational performance.
A single frontal reference fundamentally underspecifies these requirements, as it provides the model with no direct observation of how the character looks from other angles or in other expressive states.

\begin{figure}[htb]
    \centering
    \includegraphics[width=0.95\textwidth]{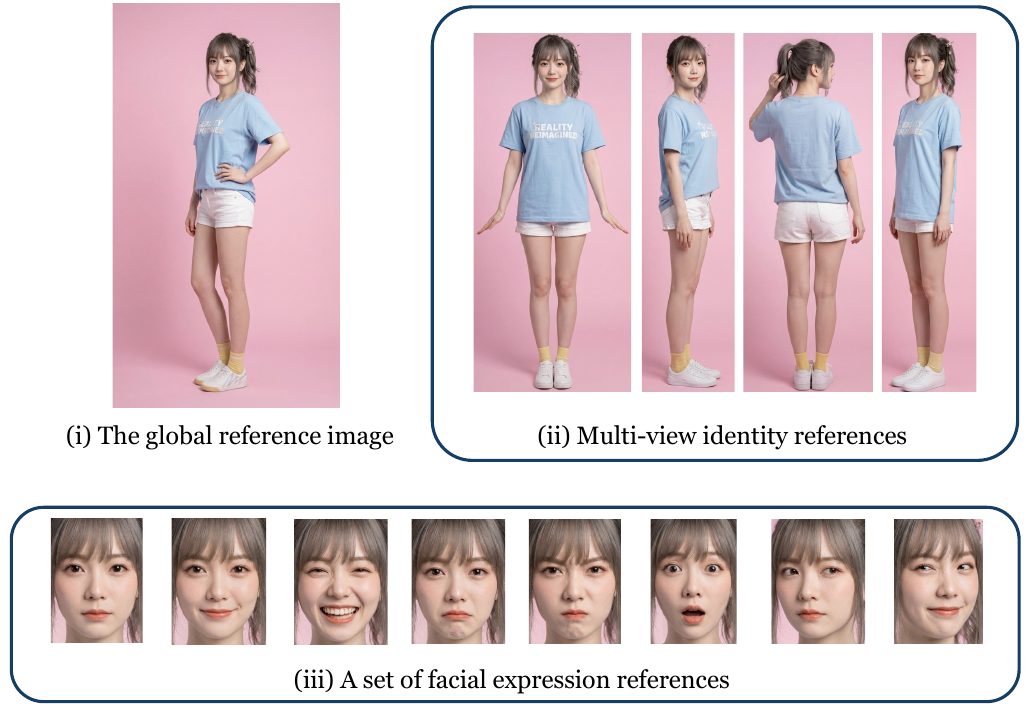}
      \caption{\textbf{An example of our proposed multi-granularity identity-aware reference images.}
      For each subject, we extract three complementary reference types from raw videos:
      \textbf{(i)} a \emph{global appearance reference} capturing overall identity and background context;
      \textbf{(ii)} \emph{multi-view body references} covering one to four viewpoints to provide appearance evidence;
      \textbf{(iii)} a set of \emph{facial expression references} spanning one to eight expressive states, enabling faithful reproduction of identity-specific details.}    
      \label{fig:refimg}
\end{figure}

To address this, we construct a set of multi-granularity reference images (see Figure~\ref{fig:refimg}) that include three complementary types of identity specification for each subject:
\emph{(1) global appearance references} that capture overall identity within the scene context;
\emph{(2) multi-view body references} that provide appearance information from different viewpoints; and
\emph{(3) facial expression references} that span a representative range of the subject's expressive repertoire.
Together, these references provide the model with a compact yet sufficient identity prior that decouples \emph{appearance} from \emph{motion}: the model can faithfully reconstruct what a character looks like from any angle and in any expression, without hallucinating details unsupported by evidence.

\paragraph{Global Appearance References.}
\label{sec:data_id_global}

The global reference anchors the character's overall visual identity and scene context, providing a holistic view of the person's appearance, attire, and surrounding environment.
A key design consideration is that if the reference frame is drawn exclusively from the same short training clip, the model may learn to copy pixels directly rather than to understand and reconstruct the character's appearance, leading to copy-paste artifacts in generation.
To prevent this shortcut, we randomly sample several candidate frames from each long-shot source video, including frames both within and beyond the short training clips, ensuring that the global reference is temporally diverse and cannot be trivially matched to any specific training frame.

\paragraph{Multi-view Body References.}
\label{sec:data_id_multiview}

To enable viewpoint-consistent generation, the model needs to know what the character looks like from angles not visible in the primary reference, such as the back of the head, side profile, or clothing details from the rear.
Without such references, the model must hallucinate these details, producing plausible but inconsistent results each time the viewpoint changes.
The challenge is to automatically determine the relative viewpoint between the camera and the person in each frame, so that we can select a diverse set of views systematically.
We address this by leveraging GVHMR~\cite{shen2024gvhmr}, a world-grounded human motion recovery method based on the SMPL 3D body model~\cite{15tog_smpl}.
GVHMR estimates the facing direction of the human body, represented by the Z-axis orientation of the SMPL root joint; we additionally estimate the camera pose using SLAM~\cite{teed2021droid}.
By computing the angle $\theta$ between the camera orientation and the human facing direction, we classify each frame into one of four viewpoint categories based on empirically determined thresholds:
frontal ($\theta \geq 155^\circ$),
rear ($\theta \leq 45^\circ$),
left-profile ($45^\circ < \theta < 155^\circ$, negative x-component of the relative direction), or
right-profile ($45^\circ < \theta < 155^\circ$, positive x-component).
From each category, representative frames are selected to compose the multi-view body reference set, providing the model with direct visual evidence for viewpoint-dependent appearance.

\paragraph{Facial Expression References.}
\label{sec:data_id_expression}

In long conversational sequences, a character's identity must remain consistent not only across viewpoints but also across expressions.
Identity-specific expression details, such as the precise shape of a character's smile, how teeth are revealed during laughter, or the pattern of wrinkles during a frown, are difficult to infer from a single neutral-expression reference and are prone to hallucination if left unspecified.
Providing a diverse set of expression references allows the model to learn these details from real observations, ensuring that the same smile always looks the same.
The challenge is to automatically identify frames that capture distinct, clearly recognizable expressions with sufficient facial detail, and to ensure that expression labels are semantically accurate.
We first restrict source videos to those with native resolution $\geq$\,1080P to ensure sufficient facial clarity.
We then use EmotiEffLib~\cite{savchenko2023facial,savchenko2022classifying} to scan video clips and identify frames containing any of eight predefined expression categories, selecting clips that include at least two distinct expressions to ensure diversity within each subject's reference set.
However, the expression category assigned to each extracted frame can be ambiguous or incorrect due to subtle or transitional expressions.
To address this, we apply a secondary verification step via an image understanding model~\cite{comanici2025gemini} to assess whether the predicted expression matches the semantic content of each frame, correcting misclassified labels to the appropriate category.

%% file: 2_base_model/1_base_architecture.tex
\section{Base LPM}
\label{sec:model}

\subsection{Base Model Architecture}
\label{sec:basic_model_arch}

\begin{figure}[t!]
    \centering
    \includegraphics[width=1\textwidth]{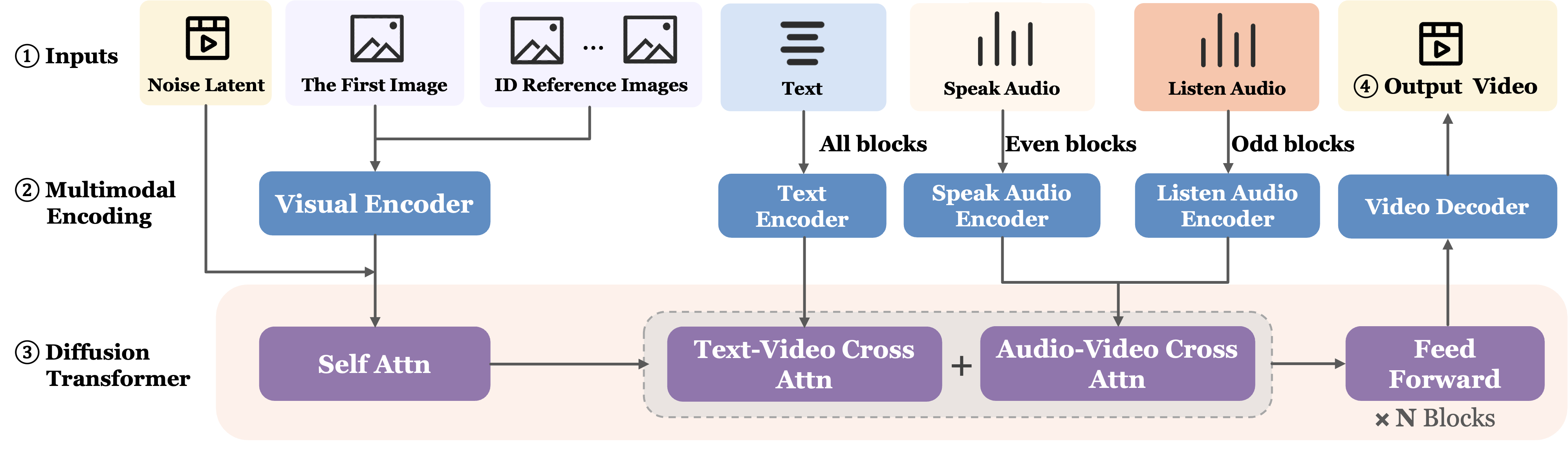}
    \caption{Base LPM architecture. Inputs (noise video, the first frame, identity-aware reference images, text, speak audio, and listen audio) are encoded by modality-specific encoders and injected into a stack of DiT blocks via self-attention (visual tokens) and cross-attention (text and audio embeddings). The output video latent is decoded by a VAE decoder to produce the generated video.}  
    \label{fig:offline_model}
\end{figure}

As shown in Figure~\ref{fig:offline_model}, we build our model upon a simple yet effective DiT architecture~\cite{peebles2023scalable}, operating on continuous patchified video latent representations.
Given a noisy video latent $x_t \in \mathbb{R}^{B \times C \times T \times H \times W}$, a diffusion timestep~$t$, text description~$c_{\text{text}}$, speak audio~$c_{\text{speak}}$, listen audio~$c_{\text{listen}}$, and a set of identity reference images $\{I_k\}_{k=1}^{K}$, the model predicts the noise $\epsilon_\theta(x_t,\, t,\, c_{\text{text}},\, c_{\text{speak}},\, c_{\text{listen}},\, \{I_k\})$.
Each DiT block follows a three-stage design:
\textbf{(1)}~self-attention with Adaptive Layer Normalization (AdaLN) for spatio-temporal modeling among noise video latent, the first image tokens, and multi-reference image tokens,
\textbf{(2)}~multi-modal cross-attention for injecting \textbf{text, speak audio, and listen audio} conditions, and
\textbf{(3)}~a feed-forward network (FFN) with AdaLN.
Text conditions are injected via standard global cross-attention.
To enable \emph{full-duplex} conversational video generation, where the character simultaneously speaks and listens, we propose an \textbf{interleaved dual-audio injection} strategy that alternately processes speak and listen audio on even and odd transformer layers.
For identity preservation, we introduce \textbf{multi-identity image token injection}, which concatenates patchified reference images directly into the self-attention sequence, enabling the model to maintain implicitly 3D-consistent character identity over long video generation.

\paragraph{Conversational Audio Conditioning.}

A core challenge in conversational video generation is modeling two concurrent audio streams: the \emph{speak audio} (driving lip sync and facial expressions) and the \emph{listen audio} (eliciting responsive reactions such as nodding, eye contact, and emotional micro-expressions).
Na\"ively injecting both audio streams into every transformer block would double the cross-attention parameters and computation, while also conflating the two semantically distinct signals within the same representational subspace.

We adopt an interleaved injection strategy in which even-numbered layers are conditioned on speaking audio, while odd-numbered layers are conditioned on listening audio.
Concretely, in the cross-attention module of each block, the query is computed as: $Q = \mathrm{RMSNorm}(W_q h)$, where $h$ denotes the video tokens, while the key--value pairs are computed from the corresponding audio context:
\begin{align}
\text{Even layers:} \quad
K_s &= W_k^{\text{spk}} c_{\text{speak}}, \quad
V_s = W_v^{\text{spk}} c_{\text{speak}}, \\
\text{Odd layers:} \quad
K_l &= W_k^{\text{lis}} c_{\text{listen}}, \quad
V_l = W_v^{\text{lis}} c_{\text{listen}}.
\end{align}
The speaking and listening branches maintain separate key, value, and output projection parameters.
Together with text cross-attention, the final output is computed as:
\begin{equation}
\mathrm{out}
=
W_o^{\text{txt}} A_{\text{text}}
+
W_o^{\text{aud}} A_{\text{audio}},
\label{eq:split_proj}
\end{equation}
where \(A_{\text{text}}\) and \(A_{\text{audio}}\) denote the outputs of the text and audio cross-attention modules, respectively.
This split projection helps reduce gradient interference between the text and audio pathways.
The interleaved design is particularly important because speaking and listening audio are associated with qualitatively different motion patterns: speaking primarily induces high-frequency local movements, such as lip synchronization and rhythm-related body dynamics, whereas listening is more correlated with lower-frequency shifts in posture and facial expression, reflecting visually grounded responses.

The proposed interleaved speak and listen audio injection has the following advantages:

\begin{itemize}

\item \textbf{Motion Distribution Alignment.}
The interleaved injection naturally aligns with the distinct motion statistics of the speak and listen data: speak data exhibit significantly higher facial and body dynamics, while listen data are characterized by subtler and slower responses, such as nodding for immersive understanding and curious expressions for a personalization response.
Dedicating separate layers to each modality allows the corresponding self-attention and FFN parameters in adjacent layers to co-specialize, forming implicit ``speak-tuned'' and ``listen-tuned'' sub-networks.
The block-level separation of speak and listen prevents gradient conflicts between modalities. Their loss landscapes benefit from independent optimization paths.

\item \textbf{Parameter and Compute Efficiency.}
The interleaved audio design injects each audio modality into only half of the transformer layers, compared to the na\"ive approach of injecting both into all layers.
This yields a 50\% reduction in audio cross-attention parameters and proportional FLOPs savings, while empirically preserving controllable quality.

\end{itemize}

\paragraph{Temporally-Aligned Audio-Video Attention.}
Rather than attending globally over the full audio sequence, the speak audio branch uses a local window of size audio frames per video sequential token, making lip synchronization precise.
This ensures that each video frame attends only to its temporally relevant audio neighborhood.
In contrast, for listen audio, we use a larger window because listening reactions would reflect semantic understanding and reasoning via a subtle nod or gaze shift, responding to the user's audio spanning longer time scales.

\paragraph{Identity Multi-reference Conditioning.}

Maintaining consistent character identity across long video generation is fundamentally challenging: the model must reproduce fine-grained facial features, hairstyle, clothing, and body proportions from arbitrary viewpoints while animating natural expressions and motions.
Prior works relying on global CLIP features lose structural detail; per-frame image cross-attention is computationally prohibitive for multiple references.

We encode each reference image~$I_k$ through the same 3D patch embedding as the unified visual latent space of various identity references, producing a set of spatial tokens per reference.
These reference tokens are \emph{concatenated to the end of the video token sequence} and participate fully in self-attention, enabling bidirectional information flow between video content and identity references at the finest granularity.
To distinguish reference tokens from video tokens, we assign them dedicated positions in the 3D Rotary Position Embedding (RoPE~\cite{su2024roformer}) space.
Specifically, \emph{expression reference images} (1--8 facial expression templates) are placed at temporal offset $t_{\text{video}} + \Delta_{\text{expr}}$, while \emph{body-view reference images} (1--4 body viewpoint images) use offset $t_{\text{video}} + \Delta_{\text{view}}$, where $\Delta_{\text{expr}}$ and $\Delta_{\text{view}}$ are fixed task-specific constants.
This design allows the model to implicitly learn the semantic distinction between expression templates and structural body references, without introducing additional learnable embeddings.
Specifically, regarding the injection of multiple reference frames, we employ a unified mechanism to bridge the noise sequence and the clean reference latent. Specifically, the ground-truth video and reference frames are both encoded through the same 3D VAE during training. Then, the video latent will be added noise, and concatenated with the clean reference latent along the sequence dimension. The combined sequence is then fed into the DiT blocks for joint self-attention computation. To handle multiple reference frames with distinct content, we employ segment-wise RoPE to explicitly distinguish them along the sequence, thereby mitigating potential conflicts during the multi-reference-injected optimization. In detail, we modify the 3D RoPE by assigning distinct temporal offsets to each reference, formulated as:
\begin{equation}
RoPE_{ij} = RoPE(t + o_i + so_j, h, w),
\end{equation}
where $i$ denotes the reference type, such as expression, multi-view, $j$ is the sub-type of reference frames, e.g., different expressions such as happy, sad, etc; here, $t$ corresponds to the temporal length of the video latent, $o_i$ and $so_j$ denote the base-offset and sub-offset assigned to reference frames; $h,w$ indicate the height and width of the reference latent. By binding each reference frame to a unique RoPE offset, the model learns to treat 3D RoPE as a weak condition that implicitly encodes reference identity, avoiding the conflict. This design supports \textbf{variable and optional numbers of reference images} ($1$--$8$ expressions $+$ $1$--$4$ body views), enabling:
\begin{itemize}
    \item \textbf{Multi-view consistency}. Multiple body viewpoints provide the model with geometric priors for head turns, profile shots, and over-the-shoulder angles.
    \item \textbf{Dynamic expression fidelity}. Multiple expression templates guide the range of feasible facial deformations, improving the naturalness of speech-driven animation.
    \item \textbf{Long-video stability}. Reference tokens persist throughout the full denoising process and across sliding-window inference, serving as identity anchors that prevent drift within ten minutes.
\end{itemize}

By injecting reference images as tokens within the self-attention sequence rather than a separate cross-attention branch, we achieve a parameter-free identity conditioning mechanism. 
The same self-attention weights that model video-to-video relationships simultaneously learn video-to-reference contexts, enabling efficient knowledge sharing.
Combined with RoPE-based positional encoding for distinguishing expression and body-view references, this unified framework supports flexible multi-reference inference for producing expressive-identity consistent video.

%% file: 2_base_model/2_base_training.tex
\subsection{Base Model Training}
\label{sec:basic_model_training}
Through flow matching~\cite{lipman2022flow}, we adopted a multi-stage training strategy to enable the base model to support multi-modal conditional control and maintain long-term character consistency.

\paragraph{Multimodal Condition Alignment.}
We initialize our model from the Text-and-Image-to-Video model (Wan2.1-I2V (16B)~\cite{wang2025wan}). We first train the speaking audio pathway from scratch on speaking-only and silence data, while keeping the pretrained text and self-attention weights largely intact to preserve the base model's visual generation capability.
At audio injection initialization, we zero-initialize the value-projection weights in the audio cross-attention layers to enable stable adaptation.
Once speaking conditioning is established, we train a listening audio pathway---also from zero initialization---on a balanced mixture of speak and listen data in each batch, enabling the model to generate responsive non-verbal behavior driven by a user's speech. We progressively combine speech, silence, listening, and conversation (both speech and listening audio) data to train their audio-alignment capabilities.
Text conditions are also injected via cross-attention layers. Richer and more precise text prompts improve fine-grained text controllability.
Notably, we remove the original CLIP image cross-attention blocks and channel-mask-image in self-attention for simplification. 
Throughout training, we apply classifier-free guidance (CFG)~\cite{cfg} to both text and audio conditions.

\paragraph{Identity Consistency Preservation.}
Meanwhile, we train the model to condition on multi-granularity identity reference images for identity-consistent generation.
A global appearance reference is provided for every training clip from the first stage. Since clips containing both expression and multi-view references are relatively scarce, we adopt a mixed training strategy. Multi-view body and facial expression references are included for 30\% of clips after the audio alignment stage, allowing the model to learn to leverage detailed identity cues when available. 
Through this procedure, the model could support arbitrary reference types and their combinations during inference.

\paragraph{Temporal Extension Training.}
Regarding the long-term video inference of the base model, the generation is typically performed in a chunk-wise autoregressive manner at the latent level. This introduces a training–inference mismatch in two aspects: (1) during training, the first frame of a window is a ground-truth visual signal, whereas the first frame may be a generated signal (continuation window) during inference; (2) due to the properties of the causal VAE, training sequences start from non-causal image latent, while continuation windows begin with causal video latent. To mitigate these inconsistencies, we concatenate the clean latent of a global reference with the generated sequence and include it in the DiT block computation, providing a stable global reference during both training and inference. In addition, with a certain probability, we randomly drop the first $2$-$5$ ground-truth video latents in the temporal dimension during training, encouraging the model to adapt to scenarios where the sequence begins with causal latents.

\paragraph{The Direct Preference Optimization Stage.} 

Despite strong overall generation quality, the base model exhibits two notable limitations in conversational scenarios.
First, in \emph{speaking} mode, large or rapid motions can introduce visual artifacts---most notably hand and limb distortions (\emph{e.g.}, unnatural bending, missing fingers, or implausible joint angles), physically inconsistent body configurations, and occasional degradation in lip-sync accuracy.
Second, in \emph{listening} mode, the model occasionally produces overly static outputs, generating near-frozen frames that lack the natural micro-movements and non-verbal responses expected of an attentive listener.
These issues are difficult to address through supervised training alone, as they involve perceptual quality dimensions, such as motion realism, artifact severity, and behavioral naturalness, that are better captured by human preference judgments than by reconstruction losses.
We therefore adopt Direct Preference Optimization (DPO)~\cite{dpo} to align the model with human aesthetic and natural preferences.
DPO avoids explicit reward modeling and replaces the costly online sampling required by methods such as PPO~\cite{schulman2017proximal} and GRPO~\cite{grpo} with a simple offline preference optimization objective, making it practical to apply at scale.

Our optimization targets two core tasks.
For \emph{speaking}, we aim to reduce hand and limb artifacts, improve the physical plausibility of body motion under large movements, and maintain precise lip-sync alignment.
For \emph{listening}, we place special emphasis on enhancing the naturalness and diversity of non-verbal behaviors~\cite{ki2026avatar}, encouraging the model to produce lively, responsive listening rather than static or repetitive outputs.
To construct the preference dataset, we generate a candidate pool of multiple video variations for each condition using distinct noise seeds.
These candidates undergo a rigorous multi-dimensional assessment covering the quality axes described above.
We adopt a Pareto-efficient selection strategy: a candidate is designated as ``preferred'' only if it strictly outperforms its counterparts in at least one metric without degrading performance in any other dimension, ensuring that the selected winner represents a genuine overall improvement rather than a trade-off.
The curated preference pairs are used to minimize the DPO loss.
We additionally incorporate a Flow Matching regularization term to prevent color shifting~\cite{du2025reg}.
Through this post-training strategy, the model demonstrates significant qualitative improvements on both speaking and listening tasks, producing more physically plausible speech motion with fewer hand and limb artifacts, better lip-sync consistency, and more natural, diverse listening behavior.

%% file: 2_base_model/3_base_inference.tex
\subsection{Base Model Inference}
\label{sec:base_inference}

Inference proceeds in three input-preparation steps followed by iterative video generation.
\textbf{(1) Visual identity.} Given a character description, we generate a first-frame portrait and a set of multi-granularity reference images to establish the character's visual identity.
\textbf{(2) Audio preparation.} For each generation segment, we prepare dual-stream audio: the speaking track is synthesized from the target dialogue script by a speech generation model, while the listening track is taken directly from the user's conversational audio input;
\textbf{(3) Text prompting.} We provide the character description, reference images, and audio content to a large language model to generate a per-chunk text prompt (125 frames each) that describes the expected motion, expression, and scene dynamics, with its style aligned to the caption distribution used during training.

With these inputs prepared, Base LPM inputs the reference images together with the dual-stream audio and per-chunk text prompts.
The two audio branches can be independently activated or muted, enabling flexible mode selection---speaking-only, listening-only, or full conversation---within a single model.
Although Base LPM is trained on short clips (3--8\,s), it supports long-form generation through chunk-wise continuation.
The target video is produced as a sequence of temporally overlapping chunks; each new chunk is initialized from the tail latents of its predecessor and linearly blended in the overlap region, providing smooth local transitions without re-training.
In practice, Base LPM can produce approximately 10 minutes of video without visible quality degradation.
This continuation strategy differs from the causal streaming setting of Online LPM (Section~\ref{sec:online_system}), where control signals arrive incrementally, and generation must remain stable over infinite horizons under real-time constraints.

%% file: 3_online_system/1_online_architecture.tex
\section{Online LPM}
\label{sec:online_system}

While Base LPM provides high-fidelity conversational performance offline, it is not directly deployable in interactive applications such as live conversation \cite{low2025talkingmachines}, virtual assistants, and games. As described in Section~\ref{sec:base_inference}, the base model can be extended to longer sequences through offline chunk-wise continuation at inference time. However, this strategy does not solve the online setting we target here. In interactive deployment, generation must be \emph{causal}, \emph{low-latency}, and \emph{stable over unbounded horizons}, while control signals such as audio and text arrive incrementally and may change during generation. Future context is unavailable, and small rollout errors can accumulate over time into oversaturation, identity drift, or structural instability \cite{liu2025rolling, huang2025self}.

These constraints introduce two coupled challenges. First, streaming control signals create a train–inference mismatch: the offline model conditions on globally encoded audio and text, whereas the online system only observes truncated and continually refreshed control context. Second, causal autoregressive rollout is inherently prone to compounding errors over long horizons, as small deviations from the desired video-latent trajectory accumulate over time. Our goal is therefore not merely to accelerate the offline teacher, but to preserve its controllability, visual fidelity, and long-horizon stability under causal streaming constraints.

To this end, we address the online problem in two parts. We first develop streaming-compatible audio and text controllers that reduce the mismatch between offline conditioning and online serving (Section~\ref{sec:streaming_audio}). We then distill the offline bidirectional DiT into a few-step causal autoregressive backbone–refiner architecture (Section~\ref{sec:online_model_architecture}), which stabilizes generation by separating coarse temporal anchoring from high-fidelity refinement. Figure~\ref{fig:online_structure} illustrates the overall architecture.

\subsection{Streaming Audio and Text Input}
\label{sec:streaming_audio}

In an online setting, control signals are inherently streaming: audio arrives incrementally, and text directives evolve, while future context is unavailable. This creates a train–inference mismatch relative to offline training, where the base model conditions on globally encoded control signals. At deployment time, however, control encoders observe only truncated history and must be refreshed repeatedly during rollout. As a result, streaming control features can deviate from their offline counterparts, leading to weakened controllability or local instability, especially near update boundaries.

We find that this mismatch is most critical for audio control. Unlike text, audio provides dense frame-level driving features for lip motion, co-speech dynamics, and temporal synchronization, and is therefore more sensitive to context truncation and boundary discontinuities. To address this issue, we adopt an overlap-aware chunk-wise audio encoding scheme and fine-tune the base model for streaming audio conditioning. At each step, the audio encoder processes a 3-second window consisting of 2 seconds of historical audio and 1 second of current audio, and the window advances with a stride of 1 second. The overlap provides temporal continuity across updates, reduces boundary artifacts, and keeps per-step latency bounded for real-time serving. Fine-tuning on 600K streaming-formatted samples substantially improves online stability and preserves the audio-driving capability of the base model under chunked inference.

Text control is also relevant in streaming interactive applications, where instructions may evolve during generation rather than being fixed at the outset. Compared with audio control, text conditioning is more robust to limited-history encoding, and we do not observe a need for additional fine-tuning. We therefore update text prompts incrementally during online generation, allowing the model to incorporate evolving high-level instructions while preserving the controllability learned offline. In practice, text updates are relatively infrequent, and each instruction often remains relevant over a longer temporal span. As a result, text control rarely becomes the bottleneck in real-time deployment; the main challenge is instead how to synchronize newly arrived instructions with ongoing generation while maintaining coherence.

\subsection{Online Model Architecture}
\label{sec:online_model_architecture}

Section~\ref{sec:streaming_audio} addresses the train--inference mismatch introduced by streaming control signals. A second challenge arises on the generation side. In offline generation, each latent update is predicted under a teacher-defined context. In online deployment, by contrast, generation must proceed causally: each new video chunk is generated using only past context, and that context is itself progressively shaped by the model's own previous predictions. The core difficulty is therefore not causal computation alone, but \emph{causal generation under self-induced history shift}. As rollout proceeds, even small local errors can perturb the future conditioning context, accumulate over time, and gradually move generation away from a valid teacher trajectory. In this sense, online few-step video distillation can be viewed as an \emph{extreme sparse-target hitting problem} over autoregressive latent trajectories: the student must reach a sharp final teacher target in only a few denoising steps while rolling out under its own imperfect history. This challenge is especially severe in the few-step regime, where the model has only a limited opportunity to correct rollout errors once they appear. The difficulty is further amplified by the offline teacher used for distillation: to achieve strong few-step visual quality, the teacher relies on classifier-free guidance (CFG), which produces trajectories that are visually sharper but also narrower and less tolerant to rollout deviations. A single-stage causal student is therefore forced to satisfy two competing requirements at once: it must remain stable under imperfect self-conditioning while also matching the final clean teacher target with high precision.

Our central methodological view is that this problem should be addressed through a \emph{sequential relaxation}. Rather than directly forcing a causal student to reconstruct the sharpest teacher trajectory, we decompose online generation into two subproblems: \emph{trajectory stabilization} and \emph{detail reconstruction}. The first stage produces a latent rollout that stays within a valid neighborhood of the teacher dynamics under autoregressive self-conditioning; the second stage starts from that stabilized trajectory and recovers the final clean latent chunk together with the high-frequency facial, appearance, and motion details that determine perceptual quality. Based on this principle, we factor the online generator into a causal \textbf{backbone} DiT and a causal \textbf{refiner} DiT. The backbone performs stable temporal anchoring and conditions on \emph{noisy-history KV caches}, i.e., history KV caches computed from forward passes with noisy history latent inputs. This places the backbone under a history distribution closer to its actual rollout regime, improving robustness to accumulated errors in self-generated history. The refiner performs high-fidelity recovery and conditions on \emph{clean-history KV caches}, i.e., history KV caches computed from forward passes with clean history latent inputs. Once temporal structure has been stabilized, clean-history conditioning provides a stronger context for reconstructing sharp, temporally coherent visual details.

\begin{figure}[htb]
    \centering
    \includegraphics[width=1\textwidth]{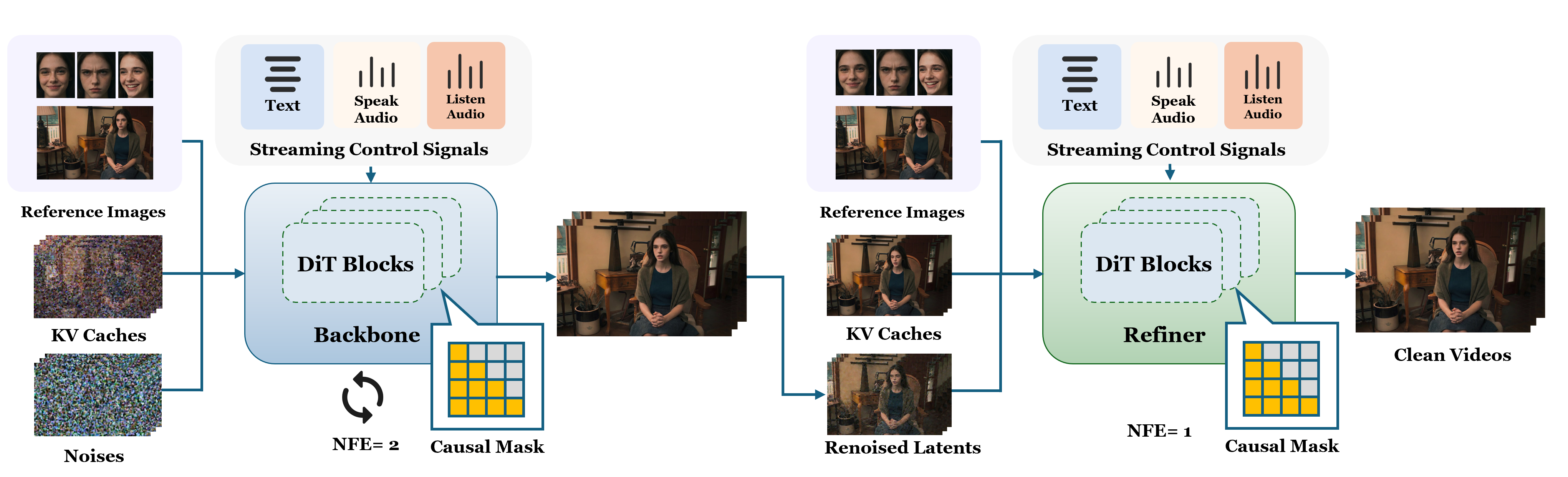}
    \caption{Online LPM architecture. The generator DiT accepts noise inputs, streaming control signals (text, speak/listen audio), and identity reference images conditioned on noisy-history KV caches to produce renoised latents. The refiner DiT then recovers the final clean video chunks conditioned on clean-history KV caches. Both stages use chunk-wise causal attention masks for autoregressive rollout.}
    \label{fig:online_structure}
\end{figure}

We instantiate this sequential relaxation within the DMD (Distribution Matching Distillation) framework~\cite{yin2024improved,yin2025causvid,yin2024onestep,poole2022dreamfusion,wang2023prolificdreamer}. As illustrated in Figure~\ref{fig:online_structure}, both the backbone and the refiner inherit the DiT architecture of the offline teacher, except that the original bidirectional attention is replaced with chunk-wise causal attention for online autoregressive generation: video latent tokens are partitioned into temporal chunks, and tokens in each chunk are allowed to attend only to the current chunk, previous video chunks, and reference-image latent tokens. In our implementation, the backbone is a distilled two-step model that maps Gaussian input noise to a coarse clean video latent trajectory conditioned on text prompts, streaming audio, identity reference images, and noisy-history KV caches. The backbone outputs are then re-perturbed with Gaussian noise and passed to a one-step causal refiner, which predicts the final clean video latent chunks conditioned on text prompts, streaming audio, identity reference images, and clean-history KV caches. Overall, this backbone--refiner decomposition matches the structure of the online generation problem itself: the backbone addresses the part dominated by exposure bias and rollout stability, while the refiner addresses the part dominated by visual precision and detail recovery, enabling stable high-fidelity long-horizon video generation in practice.

%% file: 3_online_system/2_online_training.tex
\subsection{Online Model Training}
\label{sec:online_training}

We train the online generator with a four-stage curriculum that progressively reduces optimization difficulty and narrows the gap between training and inference. The first three stages are used to train the 2NFE causal backbone. We begin with supervised ODE-based initialization on teacher denoising trajectories, then switch to off-policy DMD on teacher-derived latent states, and finally perform on-policy DMD on the backbone's own autoregressive rollouts. After the backbone has learned to maintain a valid latent trajectory, we initialize a 1NFE causal refiner from it and train the refiner with DMD to recover the final high-fidelity clean latents on top of backbone-generated trajectories.

\paragraph{DMD Objective.}
Stages~2--4 are all trained under the DMD (Distribution Matching Distillation) framework. We use $\mathcal{L}_{\text{DMD}}$ to denote the standard DMD objective~\cite{yin2024improved,yin2025causvid,yin2024onestep,wang2023prolificdreamer,poole2022dreamfusion}, which minimizes a divergence between the student distribution and the teacher-induced clean latent distribution under the corresponding training input regime. In practice, its gradient is implemented through score supervision on perturbed latent samples, given by the difference between a teacher-side real score and a dynamically estimated fake score. We adopt the standard DMD formulation, across Stages~2--4, the DMD objective is shared, while the source of the training latent states differs.

To make this distinction precise, let $p_{\theta}(\mathbf{x}^{\mathbf{t}})$ denote the distribution over chunk-wise latent sequences induced by autoregressive rollout of the current backbone model $G_{\text{backbone}}(\cdot;\theta)$ under timestep schedule $\mathbf{t}$, and let $q(\mathbf{x}^{\mathbf{t}})$ denote the distribution from which training input sequences are sampled. We call a training stage \emph{on-policy} if $q = p_{\theta}$, i.e., if the training inputs are sampled from the current model's own rollout distribution. We call it \emph{off-policy} if $q \neq p_{\theta}$; in our case, $q$ is a teacher-derived distribution constructed from offline teacher trajectories. Thus, the off/on-policy distinction refers to whether the training input distribution matches the rollout distribution induced by the current backbone.

\paragraph{Stage 1: ODE-based Initialization.}
Directly optimizing a causal student with the DMD objective is often unstable, especially at the beginning of training when the student has not yet learned a reasonable denoising trajectory. Following~\cite{yin2025causvid}, we first warm up the backbone with supervised regression on teacher denoising trajectories. This stage teaches the causal backbone to follow the online chunk-wise denoising schedule and provides a stable initialization before distribution matching.

To construct the supervision data, we sample $N$ Gaussian noise sequences $\{[x_{i,1}^{T_0}, x_{i,2}^{T_0}, \dots, x_{i,L}^{T_0}]\}_{i=1}^{N}$, where $L$ is the number of video latent chunks, and run the pretrained bidirectional teacher to obtain denoising trajectories $\{[x_{i,1}^{t}, x_{i,2}^{t}, \dots, x_{i,L}^{t}]\}_{i=1}^{N}$ at timesteps $t \in \{T_0, T_1, T_2, 0\}$ matching the online inference schedule. Let $\mathbf{t}=[t_1,t_2,\dots,t_L]$ denote a non-decreasing chunk-wise timestep schedule, where $t_\ell$ is the denoising timestep associated with chunk $\ell$. For the $i$-th sample, we define the backbone input as $\mathbf{x}_i^{\mathbf{t}}=[x_{i,1}^{t_1}, x_{i,2}^{t_2}, \dots, x_{i,L}^{t_L}]$, and the clean target as $\mathbf{x}_i^0=[x_{i,1}^{0}, x_{i,2}^{0}, \dots, x_{i,L}^{0}]$. To match the backbone's two-step generation schedule, we restrict $t_\ell \in \{T_0, T_1\}$ and train the backbone to predict the clean target:
\begin{equation}
\mathcal{L}_{\text{reg}}
=
\mathbb{E}_{i,\mathbf{t}}
\left\|
G_{\text{backbone}}(\mathbf{x}_i^{\mathbf{t}}, \mathbf{t})
-
\mathbf{x}_i^{0}
\right\|_2^2.
\end{equation}

\paragraph{Stage 2: Off-policy DMD.}
After supervised warm-starting, we transition from trajectory regression to distribution matching. In this stage, the backbone is trained with the DMD objective on teacher-derived latent states rather than on its own rollouts. Concretely, we take clean teacher latents from the ODE trajectory dataset and re-noise them under a non-decreasing chunk-wise timestep schedule with $t_\ell \in \{T_0, T_1\}$ to construct the input latents $\hat{\mathbf{x}}_i^{\mathbf{t}}$. This stage is \emph{off-policy} under the definition above, because the training latent sequences are sampled from a teacher-derived distribution rather than from the rollout distribution induced by the current backbone.

This stage serves as a stable bridge between supervised warm-starting and true rollout training: it already optimizes the backbone with the DMD objective, but still keeps the training states close to the teacher manifold. Consistent with the backbone's role in our sequential relaxation, the model is trained under noisy-history conditioning. In practice, we observe that DMD alone may lead to mode collapse. We therefore add an LPIPS perceptual regularization term~\cite{zhang2018perceptual} between the student output and the corresponding clean teacher target $\mathbf{x}_i^0$. The resulting objective is formulated as:
\begin{equation}
\mathcal{L}_{\text{off-policy}}
=
\mathbb{E}_{i,\mathbf{t}}
\left[
\mathcal{L}_{\text{DMD}}\left(G_{\text{backbone}}(\hat{\mathbf{x}}_i^{\mathbf{t}}, \mathbf{t})\right)
+
w\,\mathcal{L}_{\text{LPIPS}}\left(G_{\text{backbone}}(\hat{\mathbf{x}}_i^{\mathbf{t}}, \mathbf{t}), \mathbf{x}_i^{0}\right)
\right],
\end{equation}
where $\hat{\mathbf{x}}_i^{\mathbf{t}}$ is the re-noised latent input derived from the clean teacher target $\mathbf{x}_i^0$, and $w$ is the weight.

\paragraph{Stage 3: On-policy DMD.}
Although off-policy training is more stable, it does not fully match inference, where the backbone must condition on its own generated history. We therefore add a third stage that performs DMD on the backbone's own autoregressive rollout distribution. This stage is \emph{on-policy}, because the training latent sequences are sampled from the rollout distribution induced by the current backbone itself. It explicitly teaches the backbone to recover from its own prediction errors and is critical for long-horizon robustness.

Let $\bar{\mathbf{x}}_i$ denote the latent sequence generated by the current backbone under autoregressive rollout, and $\bar{\mathbf{x}}_i^{\mathbf{t}}$ be the corresponding re-noised input under the same chunk-wise timestep schedule $\mathbf{t}=[t_1,\dots,t_L]$. The backbone is trained under noisy-history conditioning, optimized through:
\begin{equation}
\mathcal{L}_{\text{on-policy}}
=
\mathbb{E}_{i,\mathbf{t}}
\left[
\mathcal{L}_{\text{DMD}}\left(G_{\text{backbone}}(\bar{\mathbf{x}}_i^{\mathbf{t}}, \mathbf{t})\right)
\right].
\end{equation}
Compared with Stage~2, the key change is not the loss form but the source of the training states: they are now sampled from the backbone's own rollout distribution rather than from teacher-derived trajectories, thereby substantially reducing the train--test mismatch.

\paragraph{Stage 4: Refinement DMD.}
Once the backbone has learned to stably maintain a valid latent trajectory, we train a one-step causal refiner with a refinement-oriented DMD objective to improve visual fidelity on top of backbone-generated outputs. This stage corresponds to the second half of our sequential relaxation: the backbone handles stable trajectory maintenance in an easier regime, while the refiner specializes in recovering the final clean target from that stabilized trajectory.

Specifically, we first run the trained backbone autoregressively to obtain backbone-generated latent sequences $\bar{\mathbf{x}}_i$. We then re-noise these outputs to timestep $T_2$ to construct the refiner input $\bar{\mathbf{x}}_i^{T_2}$. The refiner takes $\bar{\mathbf{x}}_i^{T_2}$ as input, together with audio, text prompts, and clean-history KV caches, and predicts the final clean latent sequence. The DMD supervision is defined with respect to the corresponding clean teacher target $\mathbf{x}_i^0$ from the ODE trajectory dataset, while the refiner inputs are sampled from the backbone-generated rollout distribution. We train the refiner with
\begin{equation}
\mathcal{L}_{\text{refiner}}
=
\mathbb{E}_{i}
\left[
\mathcal{L}_{\text{DMD}}\left(G_{\text{refiner}}(\bar{\mathbf{x}}_i^{T_2}, T_2)\right)
\right].
\end{equation}
Here $G_{\text{refiner}}$ denotes the one-step causal refiner. By training on re-noised backbone outputs rather than teacher trajectories, the refiner learns to correct residual backbone-induced artifacts and recover high-frequency details under realistic online errors.

%% file: 3_online_system/3_online_inference.tex
\subsection{Online Model Inference}
At inference time, we perform online autoregressive generation with a sliding-window decoding scheme, enabling real-time video synthesis over long and potentially unbounded horizons under bounded memory and computation. Instead of attending to the entire generation history at every step, the model only processes a fixed-size temporal window that includes the current chunk, a limited number of recent video chunks, and reference-image latent tokens. This preserves causal generation while keeping the per-step latency stable as the sequence grows. Although training is performed on 5-second clips only, the inference horizon is not inherently limited to this duration. Because generation proceeds causally in a chunk-wise autoregressive manner, the model can be rolled out beyond the training horizon to produce longer videos.

To further improve efficiency, we cache the historical key/value states before rotary positional embedding (RoPE) is applied. For each new sliding window, we dynamically apply the RoPE transformation to the cached KV states according to their updated relative positions within the current window. This avoids recomputing transformer activations over the full history, while preserving positional consistency across windows during long-horizon rollout.

Following the practice of streaming LLM inference~\cite{xiao2023efficient}, we additionally retain a small set of sink tokens in every sliding window as persistent attention anchors. These sink tokens provide a stable pathway for attention across window shifts and help reduce temporal jitter and drifting artifacts over long sequences. Together, sliding-window decoding, pre-RoPE KV caching with dynamic positional updates, and sink-token retention enable efficient and stable online generation for real-time character performance over extended durations.

%% file: 3.5_infra/1_training_optimization.tex
\section{Infrastructure}

\subsection{Training Optimization}
\label{sec:training-infra}

We developed our training infrastructure on top of TorchTitan~\cite{torchtitan} to leverage its native PyTorch distributed primitives and architectural flexibility. To support the specific demands of our architecture, we extended this foundation with a custom multimodal data pipeline, hybrid sequence parallelism, optimized activation memory management, and hardware-aware attention kernels.

\paragraph{DataLoader and Runtime Load Balancing.}

Video training clips span a wide range of lengths due to varying combinations of resolution and duration. If left unhandled, this variability leads to substantial computational skew across GPUs. To mitigate this, we precompute VAE, audio, and text embeddings offline. As a result, the data loader only needs to deliver compact latent tensors, eliminating repeated encoder passes from the active training loop. 
To further improve throughput, an online load balancer assigns incoming samples to token buckets. This equalizes the workload across data-parallel ranks and reduces padding overhead, while still preserving the randomness in sampling needed for stable gradient accumulation.

\paragraph{Parallelism Strategy.}

Our setup uses PyTorch FSDP v2 with per-block wrapping to distribute model weights, selecting sharding strategies that maximize the overlap between communication and computation. As we scale to larger clusters, we switch to Hybrid FSDP (HSDP), which shards weights within each node while replicating them across groups of nodes to reduce inter-node bandwidth.

High-resolution video modeling often generates sequences longer than 50K tokens, making full-sequence attention both computationally and memory-prohibitive. To overcome this, we use Ulysses-style context parallelism~\cite{jacobs2023ulysses}, which relies on an all-to-all collective to convert sharding along the sequence dimension into sharding along the head dimension. This rearrangement allows each rank to perform attention over the \textit{entire} sequence, but only for a designated subset of heads. We prefer Ulysses to Ring Attention mainly because our model performs extensive token manipulations along the sequence dimension; Ring-style sequential partitioning leads to substantial load imbalance across ranks and complicates these frame-dependent operations, whereas the Ulysses all-to-all transform remains invariant to how the sequence dimension is processed. For cross-attention, we restrict context parallelism to the query tensor only, since the key and value context lengths are much shorter.

\paragraph{Memory-Efficient Activation Management.}

To reduce GPU memory usage without paying the cost of recomputing entire layers, we use operator-level selective activation checkpointing (SAC). We checkpoint only the computationally intensive operations—such as matrix multiplications, attention kernels, and collectives used in context parallelism—while dropping inexpensive element-wise intermediate results.
Under tight memory budgets, we further combine SAC with asynchronous CPU offloading. By transferring a configurable subset of checkpointed tensors to CPU memory, we exchange PCIe bandwidth for critical GPU memory savings, which in turn supports larger batch sizes and extended context lengths.

\paragraph{Efficient Attention and Kernel Optimizations.}

We expose attention operations through a single unified interface that automatically selects the most efficient backend available, be it FlashAttention-2~\cite{dao2023flashattention2}, FlashAttention-3~\cite{shah2024flashattention3}, or CuTe-based kernels~\cite{zadouri2026flashattention}. To avoid padding overhead in heterogeneous batches, we use variable-length (\texttt{varlen}) implementations that pack multiple sequences into a single kernel launch.
We further implement memory-bound components such as RMSNorm, activation functions, and RoPE as custom Triton kernels specialized to our model’s tensor layouts. This fusion ensures that intermediate results remain in fast registers instead of being written out to high-bandwidth memory. 
For autoregressive model variants (Section~\ref{sec:online_system}), we employ PyTorch FlexAttention~\cite{dong2024flexattention} together with \texttt{torch.compile}. This setup JIT-compiles a block-sparse causal mask, strictly preserving temporal causality while achieving dense-kernel throughput, and avoids ever materializing the full $N \times N$ mask.

\color{black}  

%% file: 3.5_infra/2_online_inference_optimization.tex
\subsection{Online Interactive Video System}
\label{sec:online_model_system}
Online LPM serves as the core video generation engine for a broad class of \emph{interactive video systems}, including conversational agents or avatars, live streaming characters, and game NPCs. Although these applications differ in interaction semantics, they share a common systems requirement: the model must operate as a persistent, stateful online service that continuously produces temporally coherent video while remaining responsive to streaming user inputs and control updates.

From a systems perspective, the online interactive video system takes as input a character identity specification together with streaming conditioning signals, such as speaking/listening audio, text, or other external control inputs, and produces a continuous video stream with synchronized visual performance. To support practical deployment, such a system should satisfy three key requirements: (1) \emph{low-latency continuous generation}, so that visual output can be streamed in real time; (2) \emph{state-adaptive responsiveness}, so that the character can react promptly to changing inputs, interruptions, or interaction state transitions; and (3) \emph{long-horizon consistency}, so that identity, motion, and temporal coherence are preserved over extended sessions with bounded memory cost.

\begin{figure}[htbp]
\centering
\includegraphics[width=\textwidth]{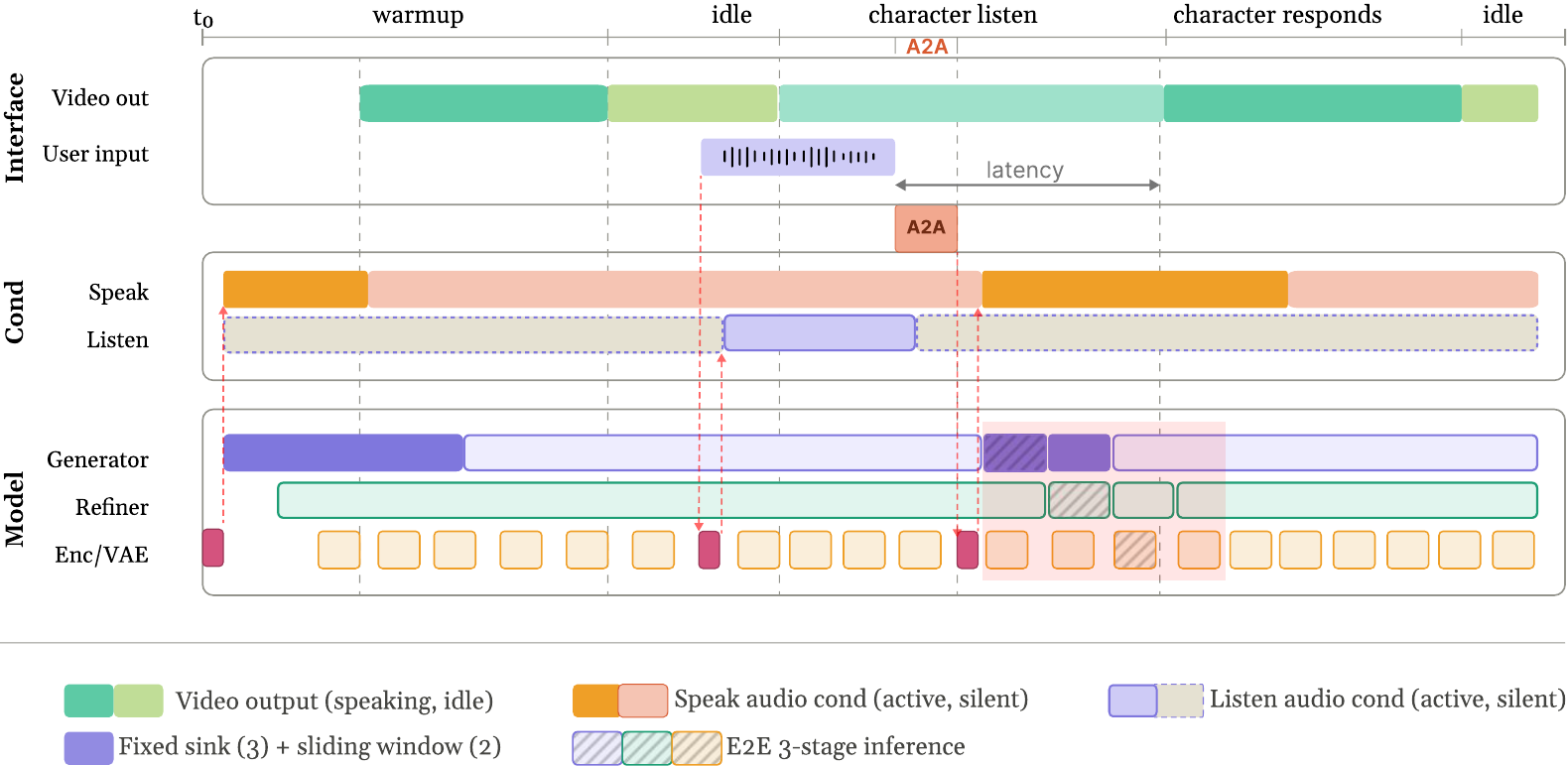}
\caption{\textbf{Execution timeline of the online interactive video system, illustrated with a full-duplex dialogue example.} The system progresses through warmup, idle, listening, and responding states while maintaining continuous video output. Audio conditions (listen/speak) are aligned with the streaming timeline, and the Audio2Audio module introduces bounded latency between user input and response. The highlighted region shows the overlapping three-stage pipeline (Generator, Refiner, VAE) operating on chunked inputs. The cache structure (fixed sink + sliding window) enables stable long-horizon generation. Text conditioning and KV-cache operations are omitted for clarity.}
\label{fig:online_system_time}
\end{figure}

To achieve these requirements, we build a persistent online runtime around Online LPM, as illustrated in Figure~\ref{fig:online_system_time}. The runtime coordinates model execution, multimodal conditioning, and session state through three architectural pillars:
\begin{itemize}
    \item \textbf{Pipelined Inference Execution:} Autoregressive generation is discretized into fixed 1-second chunks (24 fps), matching the streaming units in Figure~\ref{fig:online_system_time}. Each chunk is processed through a three-stage pipeline (Generator, Refiner, VAE) with overlapping execution, where refinement of the current chunk proceeds in parallel with generation of the next. This overlap reduces effective latency while maintaining throughput. The text/audio encoders share resources with the VAE for improved utilization. In practice, the Generator and Refiner each take $\sim$700\,ms, while the VAE requires 180\,ms, with negligible encoder overhead.

    \item \textbf{Multimodal State-Adaptive Control:} The runtime is driven by streaming multimodal signals, including identity specifications, audio/text inputs, external control signals, and cached session state. It transitions across runtime states such as \textit{warmup}, \textit{idle}, \textit{listening}, and \textit{responding} with low latency, enabling seamless adaptation to different interaction modes while preserving identity consistency and audio-visual synchronization.

    \item \textbf{Efficient Context Management:} To maintain long-horizon visual consistency with constant memory, the runtime adopts a hybrid cache with \textbf{sink tokens spanning 3 chunks} for global identity preservation and a \textbf{sliding window of 2 chunks} for local temporal continuity, as shown in Figure~\ref{fig:online_system_time}. Together with the current chunk, this forms a 5-chunk context, enabling stable and efficient autoregressive updates independent of session length.
\end{itemize}

This runtime is application-agnostic and supports a range of online interactive video settings. As one representative instantiation, it can be coupled with an external audio service, denoted as the \textit{A2A} (\emph{i.e.}, audio2audio) module, to enable live full-duplex conversational interaction. This module can be an end-to-end audio-to-audio system (\emph{e.g.}, Doubao or GPT-4o~\cite{hurst2024gpt4o}), an ASR--LLM--TTS pipeline, or an LLM-based dialogue system with TTS, as long as it provides streaming audio output. Figure~\ref{fig:online_system_time} illustrates this full-duplex dialogue workflow as a concrete example of the broader system.

\subsection{Online Inference and System Optimization}
\label{sec:online_optimization}

We optimize the deployed system at two levels: \textit{inference efficiency} for high throughput and \textit{system responsiveness} for real-time interaction, both critical for a seamless user experience.

\paragraph{Model Inference Optimization.}

To evaluate deployment readiness, we track per-chunk latency, time-to-first-response (TTFR), and steady-state throughput. Latency governs responsiveness, while throughput ensures sustained 24\,fps streaming.
The main bottleneck lies in the dual 17B DiT backbones. We optimize them with fused kernels and efficient attention implementations (e.g., FA4~\cite{zadouri2026flashattention}) via \texttt{torch.compile}, achieving $\sim$0.35\,s per chunk (1-NFE) on a single GPU, enabling real-time high-quality video generation.
The full multimodal stack (encoders, refiners, VAE) is further parallelized using pipeline execution, allowing different chunks to occupy different stages concurrently and improving utilization.

\paragraph{System-Level Optimization.}

Fluid interaction requires not only speed but also robust handling of interruptions and role transitions. We adopt the following mechanisms:
\begin{itemize}

\item \textbf{State Splitting:} We separate \textit{persistent visual state} from \textit{refreshable condition caches}, enabling rapid response to new inputs without disrupting visual continuity.

\item \textbf{Boundary-Aligned Updates:} Updates are applied at chunk boundaries (Figure~\ref{fig:online_system_time}). Each chunk completes under fixed conditions, and new inputs take effect in subsequent chunks.

\item \textbf{Controlled Lookahead:} The scheduler limits generation to stay slightly ahead of playback, reducing backlog and minimizing latency during interruptions.
\end{itemize}

To maintain coherence, updated audio representations are aligned with the active chunk timeline before submission, ensuring synchronized speech and facial motion. These optimizations enable stable long-horizon generation with bounded-delay control updates.

%% file: 5_eval_sections/1_benchmark.tex
\section{Evaluation}
\label{sec:evaluation}

\subsection{LPM-Bench: A Multimodal Human Performance Benchmark}
\label{sec:vihp_bench}

Existing video generation
benchmarks~\cite{yang2025videogen,zeng2024dawn,huang2024vbench,liu2024evalcrafter,han2025video,sun2025t2v,huang2025vbench++,zheng2025locot2v} primarily focus on general visual quality metrics or short-term text-to-video and image-to-video tasks instead of conversational performance. While useful, these inputs and metrics do not adequately assess the perceptual attributes central to human-centric video generation, such as behavioral expressiveness, conversational naturalness, and long-horizon character consistency. They are therefore insufficient for evaluating whether a generated character appears convincing and coherent over time.
To address this gap, we construct \textbf{LPM-Bench}, a comprehensive benchmark covering the full range of conversational performance capabilities. The benchmark is structured in two layers: a \emph{functional evaluation layer} that tests each conversational mode with dedicated
samples and assessment criteria, and a \emph{generalization evaluation layer} that verifies robustness across diverse identities, motions, and multimodal input conditions. In total, LPM-Bench comprises \textbf{1{,}000} test cases organized into hierarchical subsets.

\paragraph{Functional Evaluation.}
The functional layer evaluates three core conversational modes:
\begin{itemize}
  \item \textbf{Speaking} ($\sim$400 samples): audio-driven generation
  spanning \emph{78 emotions}, \emph{22 expression bases},
  \emph{co-speech gesture}, \emph{singing types} (short- and long-form
  across vocal styles), \emph{pronunciation} (especially for bilingual articulation fidelity in English and Chinese), and motions including full-body movement, gesture, interaction, and physical plausibility. 
  \item \textbf{Listening} ($\sim$200 samples): non-verbal responsive
  behavior conditioned on a partner's or user's speech, evaluated across
  varying interpersonal relationships, character personas, speech content,
  emotional context, and bilingual settings, assessing both the presence of
  attentive listening states and the semantic appropriateness of the response.
  \item \textbf{Conversation} ($\sim$200 samples): multi-turn video-audio
  conversations with natural speaking--listening transitions within various
  speech content, emotional context, and bilingual settings, ranging from
  single-turn to extended multi-turn dialogues.
\end{itemize}

\paragraph{Generalization Evaluation.}
The generalization layer probes robustness beyond the core conversational
modes:
\begin{itemize}
  \item \textbf{Diverse Human Motion} ($\sim$100 samples): evaluates
  generation quality for a broad range of human activities beyond
  conversation, including dynamic body movement, object and environmental interaction, and varied physical actions.
  \item \textbf{Character Generalization} ($\sim$100 samples): tests
  identity-consistent generation across diverse appearance styles,
  including photorealistic, anime, 3D-rendered, and artistic characters, to verify that the model generalizes beyond photorealistic human subjects.
\end{itemize}

\paragraph{Systematic Diversity.}
LPM-Bench is designed for controlled diversity across three axes:
\begin{itemize}
  \item \textbf{Appearance}: test subjects span a wide range of visible
  extents (from close-up to full-body long shot) and appearance attributes
  (artistic styles, ethnicities, ages, clothing, initial expressions, and
  poses).
  \item \textbf{Performance motion}: annotations cover 22 expression
  bases, 78 emotions, 18 gaze and tear statuses, over 5{,}000 motion
  descriptors, and 400 head pose configurations.
  \item \textbf{Audio}: test audio covers diverse voice timbres,
  paralinguistic cues (laughter, sighs, hesitation), and languages. The
  majority of test audio ranges from 5 to 180 seconds; an additional 10\%
  of the benchmark focuses on long-form generation---singing, speaking, and
  conversation at durations from 3 minutes to 1 hour---to stress-test
  temporal consistency and identity stability over extended horizons.
\end{itemize}

Across all modes, a dedicated \emph{identity-consistency} subset evaluates the proposed multi-reference conditioning, verifying that identity is
preserved without copy-paste artifacts. For external comparison, we report
on the speaking subsets; internally, all modes are used for comprehensive
evaluation.

\begin{figure}[!t]
    \centering
    \includegraphics[width=1\textwidth]{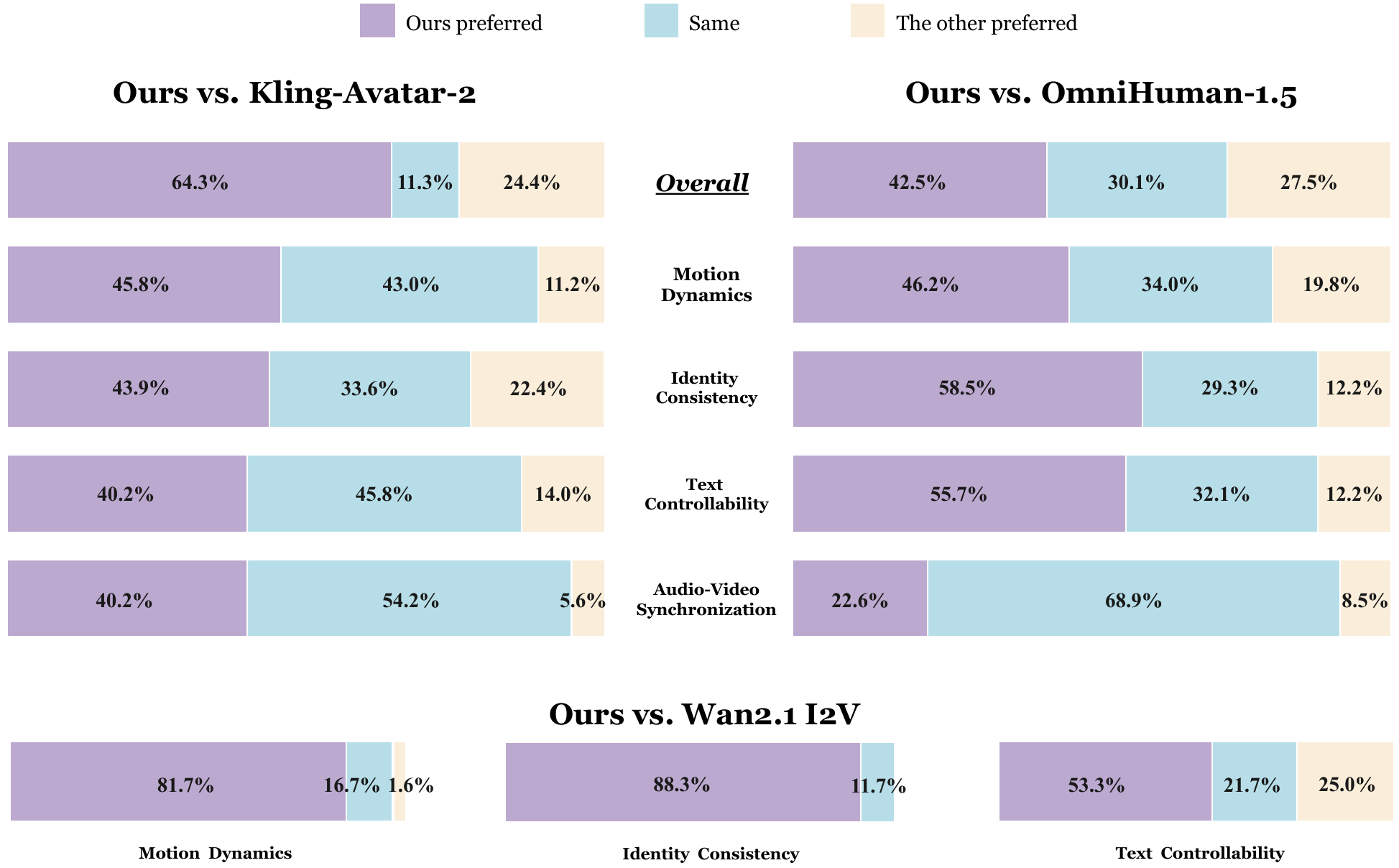}
    \caption{Human preference evaluation (G/S/B) of Base LPM (720P) versus Kling-Avatar-2, OmniHuman-1.5 and Wan2.1-I2V on \textbf{LPM-Bench}.
    \textbf{Overall} measures which video appears more like a real person. The four diagnostic dimensions evaluate pairwise preference along specific axes.}
    \label{fig:GSB_offline_model}
\end{figure}

\paragraph{Evaluation Dimensions and Metrics.}
LPM-Bench evaluates generated videos via four dimensions:
\begin{itemize}
  \item \textbf{Motion Dynamics}: temporal coherence and physical plausibility of generated motion, penalizing deformation, flickering, body-part disappearance, and collapse into near-static outputs.
  \item \textbf{Identity Consistency}: whether facial and body attributes, and overall appearance, remain perceptually consistent with the reference image throughout generation.
  \item \textbf{Text Controllability}: how faithfully the generated video follows text instructions, including specified motion, gaze, expression, emotion, and temporal ordering.
    \item \textbf{Audio-Video Synchronization}: evaluates the correspondence between the audio stream and visible behaviors across three scenarios.
    For \emph{speaking}, we assess lip-sync precision with respect to speech content, as well as the consistency between audio rhythm, emotion, and the corresponding visual motion and facial expression.
    For \emph{listening}, we measure the success rate of suppressing false lip movements when the character is not speaking, and the subjective alignment between the user's audio and the listener's visual response in terms of emotion, semantic context, interpersonal relationship, and personality.
    For \emph{conversation}, we additionally evaluate the naturalness and coherence of speak-listen transitions, including turn-taking timing, expression continuity across role switches, and consistency over multi-turn dialogue.
\end{itemize}
We adopt two complementary human-evaluation protocols: (1) A \emph{pairwise comparative study} uses the Good/Same/Bad (GSB) framework: raters compare our result with a baseline generated from the same input, each pair is judged by three independent raters, and the final label is determined by majority vote; (2) An \emph{absolute score evaluation} on a 1--5 Likert scale~\cite{likert1932technique} provides finer-grained per-sample quality judgments.
We apply pairwise GSB to all four diagnostic dimensions, plus an additional \textbf{Overall} dimension that captures holistic preference in an arena-style~\cite{chiang2024chatbot} comparison; absolute scoring is applied only to the four diagnostic dimensions.
Together, these protocols provide both relative preference and absolute quality assessment across all evaluation process.

%% file: 5_eval_sections/2_offline_compare.tex
\begin{figure}[!t]
    \centering
    \includegraphics[width=\linewidth]{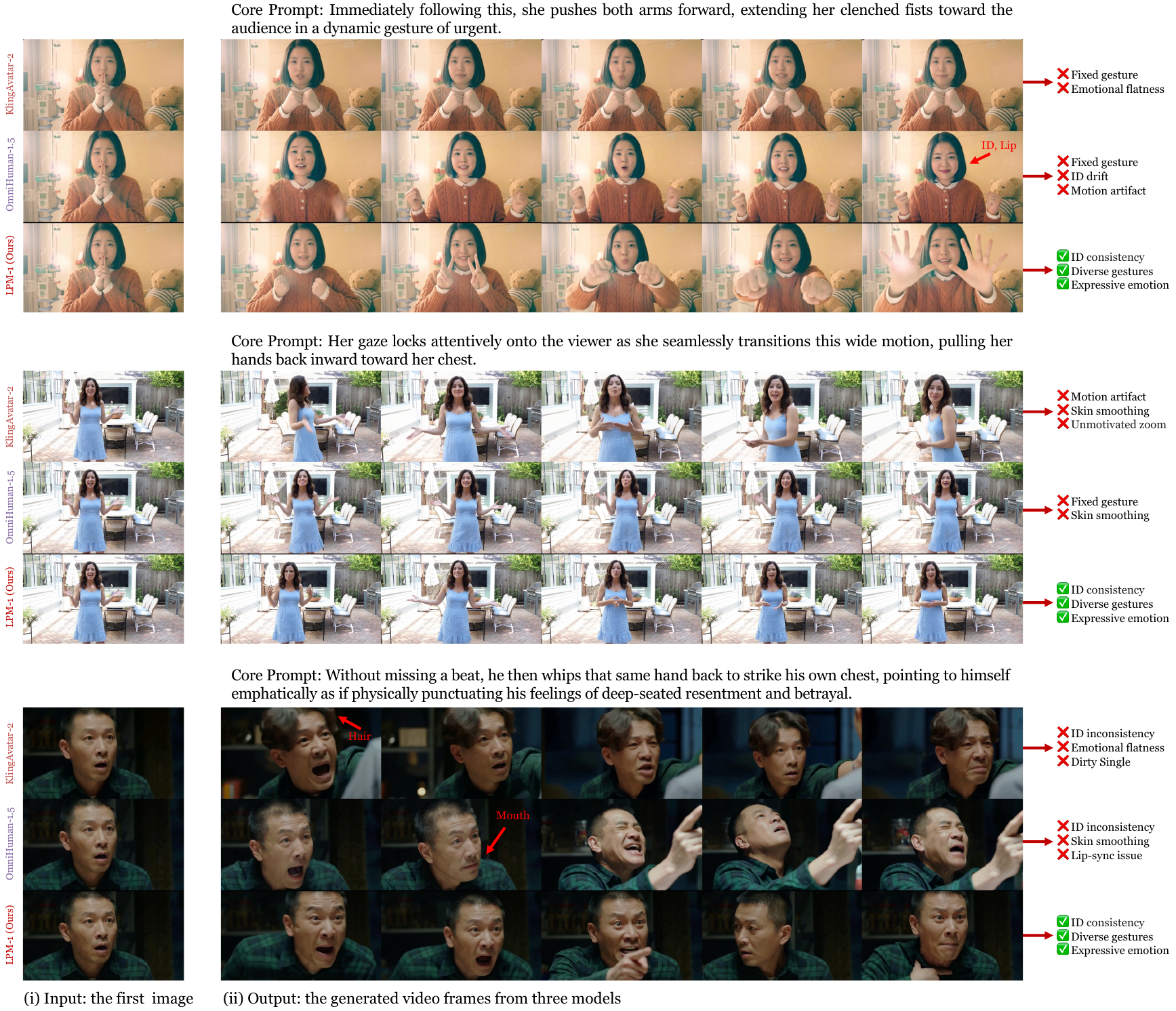}
    \caption{\textbf{Qualitative comparison of Base LPM (720P) with Kling-Avatar-2 and OmniHuman-1.5.} Each row shows sampled frames from videos generated with the same reference image, text, and audio. Our model produces more diverse gestures, expressive emotion, accurate lip synchronization, and consistent identity.}
    \label{fig:combined_base_results}
\end{figure}

\subsection{Base Model Evaluation}
\label{sec:offline_model_evaluation}

\paragraph{Summary.} Base LPM (720P) is evaluated on LPM-Bench against two state-of-the-art baselines under the same resolution: Kling-Avatar-2~\cite{team2025klingavatar} and
OmniHuman-1.5~\cite{jiang2025omnihuman}. Meanwhile, we also compare the baseline model Wan2.1-I2V (16B)~\cite{wang2025wan}. Two key advantages emerge:
\begin{itemize}[noitemsep,topsep=2pt,leftmargin=*]
    \item \textbf{Long-form live video generation.} Base LPM supports
    minute-length generation with stable visual fidelity, precise text controllability, and consistent identity preservation, whereas both Kling-Avatar-2 and OmniHuman-1.5 are limited to approximately 30 seconds.
    \item \textbf{Short-form cinematic performance.} Even within short-form clips, Base LPM delivers superior acting quality with natural, contextually appropriate performance and emotionally precise delivery.
\end{itemize}

\paragraph{Human Preference Results.}
Figure~\ref{fig:GSB_offline_model} summarizes pairwise Good/Same/Bad (G/S/B) human preference results. Base LPM is preferred overall by a
large margin against both baselines: 64.3\% vs.\ Kling-Avatar-2 (11.3\% same, 24.4\% other) and 42.5\% vs.\ OmniHuman-1.5 (30.1\% same, 27.5\% other). The strongest advantages appear on \emph{identity consistency} (58.5\% preferred) and \emph{text controllability} (55.7\% preferred),
followed by \emph{motion dynamics} (46.2\% preferred), all with approximately 30\% same and 12--20\% other preferred. These results indicate that Base LPM provides stronger holistic realism, with the most consistent gains in identity preservation, instruction following, and stable motion generation.

\begin{figure}[!t]
    \centering
    \includegraphics[width=1\textwidth]{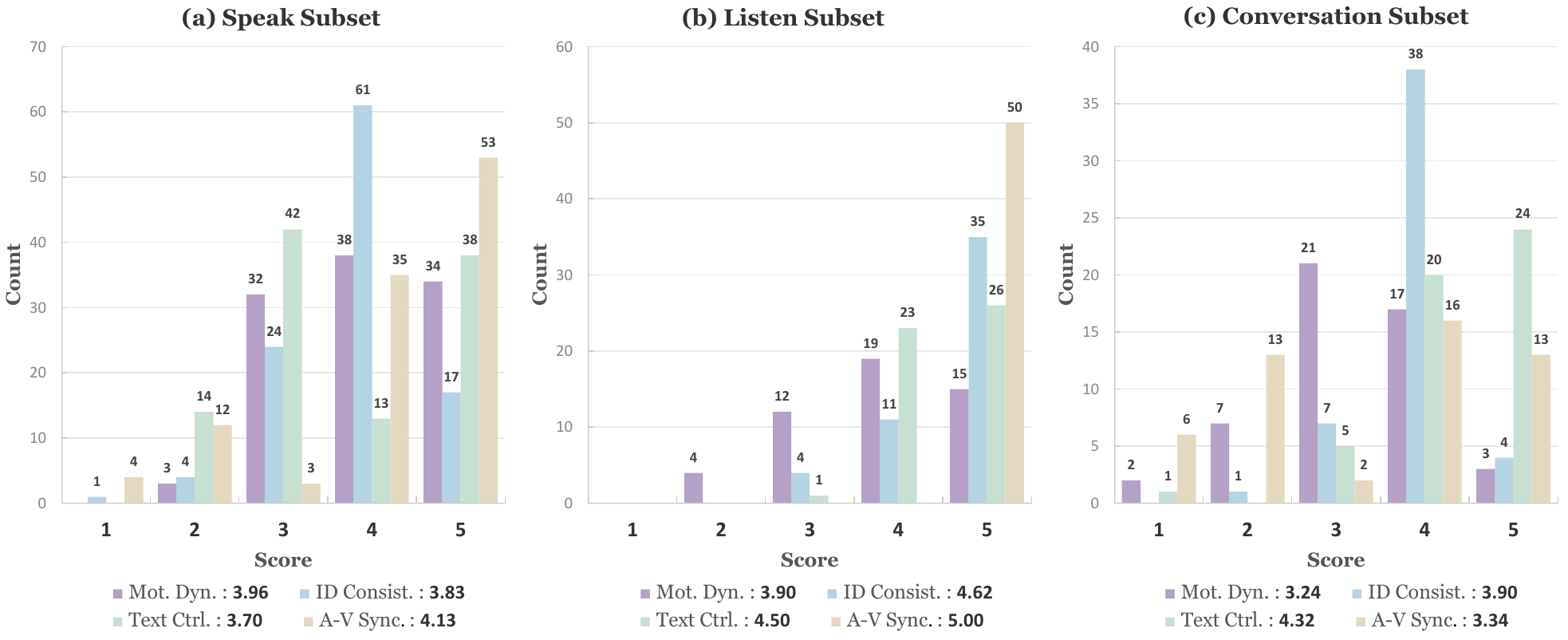}
    \caption{Absolute score distributions (1--5 Likert) of Base LPM
    (720P) on LPM-Bench across Speak, Listen, and Conversation. Legend
    shows per-dimension means.}
    \label{fig:base_model_evaluation}
\end{figure}

\paragraph{Dimension-wise Analysis.}
The preference patterns are consistent with qualitative inspection across
the four diagnostic dimensions (also see Figure~\ref{fig:combined_base_results} for representative examples):
\begin{itemize}
    \item \emph{Identity Consistency} is the clearest strength.
    Kling-Avatar-2 tends to drift toward darker, over-smoothed skin
    tones; OmniHuman-1.5 exhibits facial drift and color-tone shifts
    after head motion; Wan2.1-I2V (16B)~\cite{wang2025wan} suffers from more severe facial drift, artifacts, slow motion dynamics, worse text controllability, and appearance degradation, and it only generates 5-second videos. Base LPM retains natural skin texture and remains closer to the reference throughout the generation.
    \item \emph{Motion Dynamics and Text Controllability.} Both baselines
    frequently exhibit fixed gestures, structural collapse in hands and
    teeth, or incomplete responses to gaze and body-motion instructions.
    Base LPM follows multimodal instructions more reliably, especially
    for coordinated non-verbal behavior, and maintains more coherent body
    structure during expressive movement.
    \item \emph{Audio-Video Synchronization.} The clearest gain over
    Kling-Avatar-2 is an articulation during emotional speech, where it
    frequently under-articulates despite energetic vocal delivery.
    OmniHuman-1.5 is closer on this dimension overall, though it more
    often exhibits exaggerated lip motion or responses to background
    audio cues.
\end{itemize}

\paragraph{Absolute Quality Assessment.}

We further evaluate Base LPM using a 1--5 Likert scale across three scenarios (Figure~\ref{fig:base_model_evaluation}). \textbf{Listen} achieves the highest quality (average 4.51), with perfect audio-video synchronization (5.00) and strong identity preservation (4.62), confirming
that the model produces temporally accurate and emotionally coherent
listening responses. \textbf{Speak} scores 3.91 on average, with
audio-video synchronization as the strongest dimension (4.13, 49.5\% at
maximum score); the primary bottleneck is text controllability (3.70),
which exhibits a bimodal ``all-or-nothing'' execution pattern concentrated
on \textsc{Action} and \textsc{Gaze} failures. \textbf{Conversation}
(average 3.70) poses the greatest challenge: motion dynamics drops to 3.24
as hand articulation degrades in longer sequences, and audio-video
synchronization (3.34) is disrupted by frequent speaking--listening
transitions. Identity consistency (3.90) and text controllability (4.32)
remain comparatively robust, though 78\% of controllability failures
concentrate in multi-paragraph action sequences, identifying temporally
ordered action execution as the central remaining bottleneck.

%% file: 5_eval_sections/3_online_compare.tex
\begin{figure}[!t]
        \centering
        \includegraphics[width=1\textwidth]{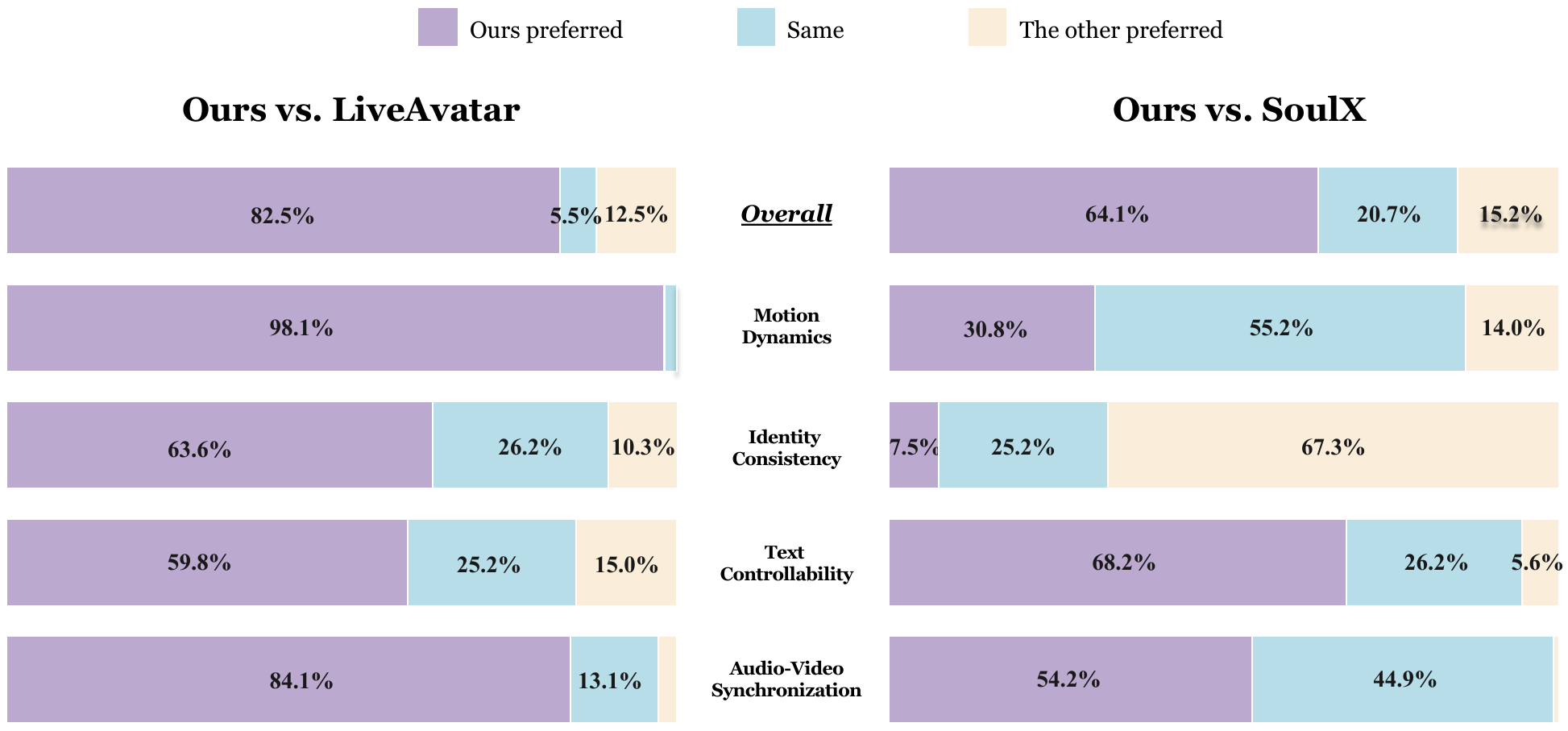}
        \caption{Human preference evaluation (G/S/B) of Online LPM (480P) versus LiveAvatar and SoulX on \textbf{LPM-Bench}. Each video pair is evaluated by three independent raters. The Overall axis uses an arena-style realism judgment, while the remaining axes separately evaluate motion dynamics, identity consistency, text controllability, and audio-video synchronization. Online LPM is strongly preferred to LiveAvatar across all dimensions, and is preferred overall to SoulX despite SoulX's advantage on identity consistency.}
        \label{fig:GSB_online_model}
    \end{figure}
    
\subsection{Online Model Evaluation}
\label{sec:online_model_evaluation}

\paragraph{Summary.} Online LPM (480P) is evaluated on LPM-Bench against two real-time
baselines: LiveAvatar~\cite{liveavatar} and SoulX~\cite{shen2025soulx}.
Two key advantages emerge:
\begin{itemize}[noitemsep,topsep=2pt,leftmargin=*]
    \item \textbf{Long-form live video generation.} Online LPM produces
    vivid performance with precise lip-sync and audio alignment, and
    natural expression matching melody and speech over indefinite
    durations.
    \item \textbf{Short-form cinematic performance.} Online LPM delivers
    richer facial micro-expressions, natural gestures in sync with audio,
    well-paced performance rhythm, cinematic-quality emotion alignment,
    and faithful scene restoration.
\end{itemize}

\begin{figure}[!t]
    \centering
    \begin{subfigure}{\textwidth}
        \centering
        \includegraphics[width=\linewidth]{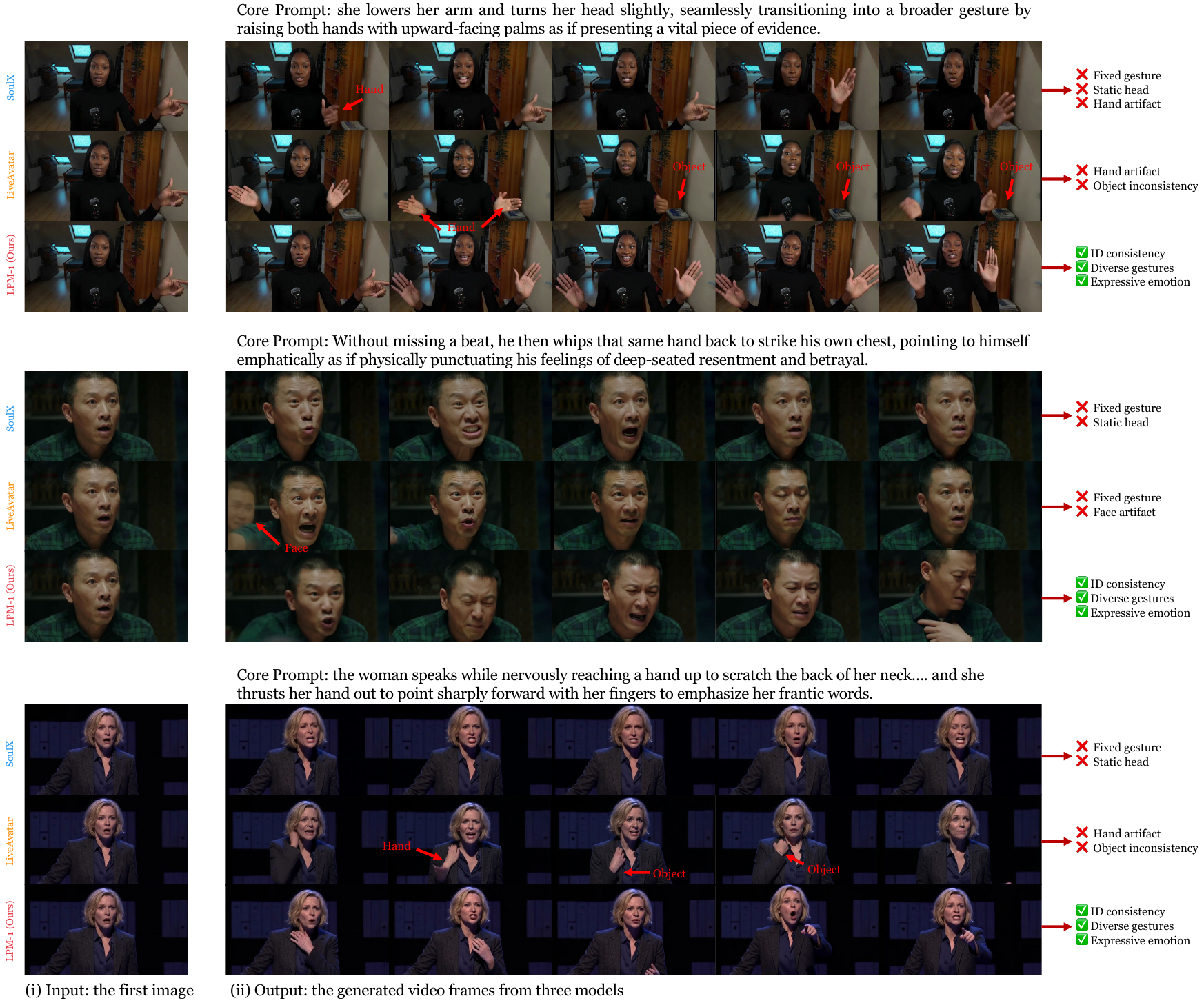}
    \end{subfigure}
    \caption{Qualitative comparison of Online LPM (480P) with SoulX and LiveAvatar. Each row shows sampled frames generated from the same reference image and audio input. SoulX tends to produce conservative motion with limited gesture and head-pose variation, while LiveAvatar exhibits structural instability, background inconsistency, and lip-sync errors. Online LPM better balances motion richness, identity preservation, and temporal coherence.}
    \label{fig:combined_online_results}
\end{figure}
    
\paragraph{Comparison with SOTA Baselines.}
Figure~\ref{fig:GSB_online_model} presents G/S/B human preference results (see Figure~\ref{fig:combined_online_results} for qualitative examples).
Online LPM is strongly preferred over LiveAvatar across all dimensions
(82.5\% preferred overall, 5.5\% same, 12.5\% other), with the largest
margins on \emph{Motion Dynamics} and \emph{Audio-Video Synchronization}.
Compared with SoulX, Online LPM is preferred overall (64.1\% preferred,
20.7\% same, 15.2\% other) and favored on \emph{Motion Dynamics},
\emph{Text Controllability}, and \emph{Audio-Video Synchronization}, while
SoulX is preferred on \emph{Identity Consistency}. This pattern reflects a
trade-off between conservative portrait stability and behaviorally rich
generation: SoulX often produces near-frontal faces with limited motion
amplitude, which aids identity judgments but reduces perceived liveliness.
Under the \emph{Overall} criterion, which directly asks which video looks
more like a real person, raters consistently weigh behavioral realism over conservative appearance preservation.

\begin{figure}[!t]
    \centering
    \includegraphics[width=1\textwidth]{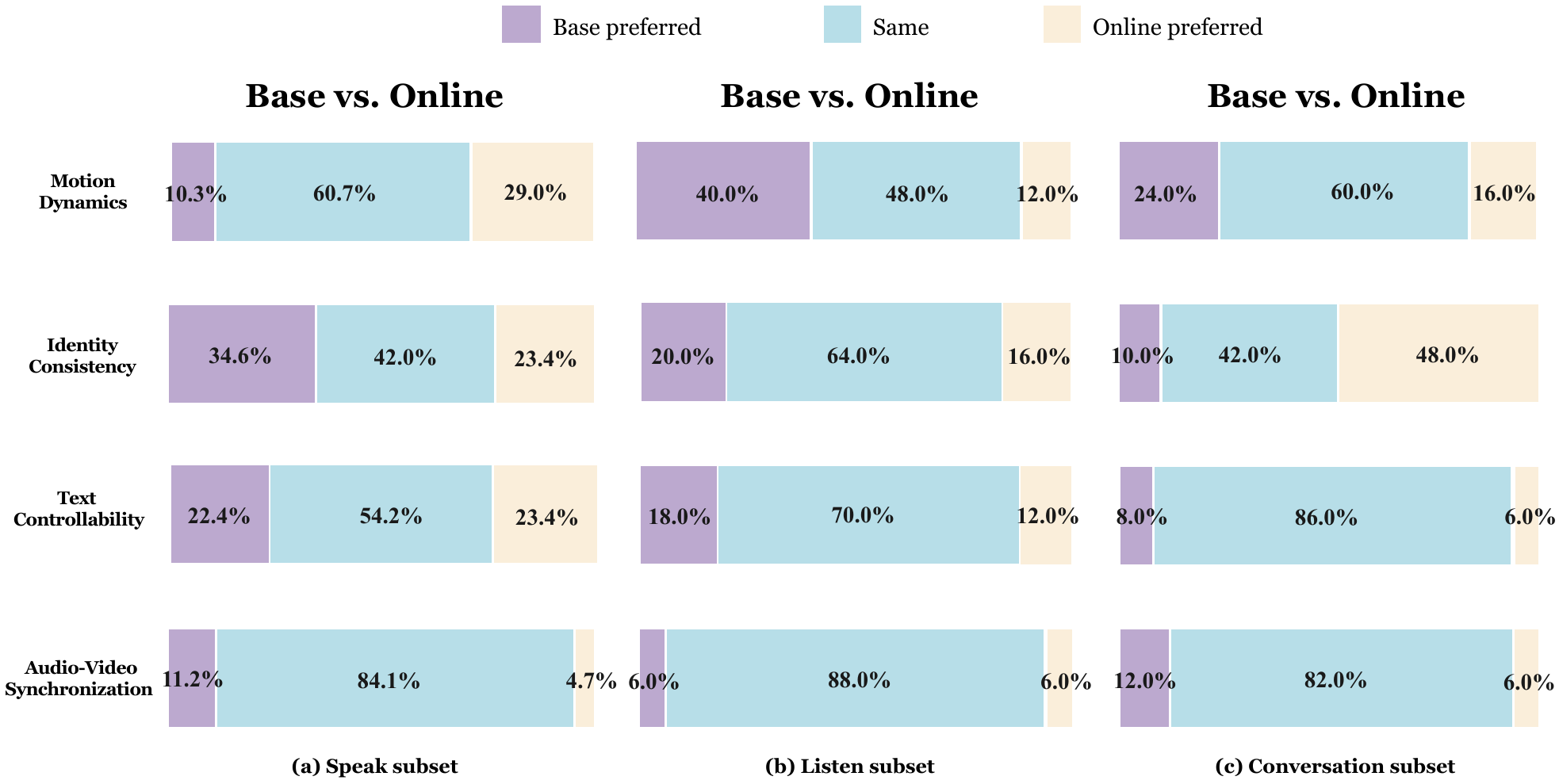}
    \caption{Human preference evaluation (G/S/B) of Base LPM (480P) and Online LPM (480P) across three performance scenarios: (a) Speak, (b) Listen, and (c) Conversation. The online model remains competitive in terms of motion dynamics and identity consistency, while text controllability and audio-video synchronization aspects are similar.}
    \label{fig:GSB_offline_and_online_model}
\end{figure}
    
\paragraph{Online LPM vs.\ Base LPM.}
To characterize quality trade-offs from real-time distillation, we compare
Online LPM with Base LPM across three interaction subsets at 480P
(Figure~\ref{fig:GSB_offline_and_online_model}):
\begin{itemize}
    \item \emph{Speak}: The two models are broadly competitive, with high
    Same rates across all dimensions (54--84\%). Online LPM is favored on
    motion dynamics (29.0\% vs.\ 10.3\% Base), while Base LPM is stronger
    on identity consistency (34.6\% vs.\ 23.4\% Online). Text
    controllability is evenly split (22.4\% vs.\ 23.4\%), and audio-video
    synchronization is overwhelmingly judged as equivalent (84.1\% Same),
    confirming that distillation preserves core speech-conditioned
    capabilities.
    \item \emph{Listen}: Base LPM is preferred, primarily driven by motion
    dynamics (40.0\% Base vs.\ 12.0\% Online). This scenario depends on
    subtle, low-amplitude reactive behavior that the online model's
    temporal regularization tends to suppress. Other dimensions show high
    agreement (64--88\% Same), and audio-video synchronization is
    perfectly tied (6.0\% each). This remains the main quality gap between
    the two systems.
    \item \emph{Conversation}: Online LPM shows its clearest advantage on
    identity consistency (48.0\% Online vs.\ 10.0\% Base), confirming
    superior robustness over long multi-turn sequences. On-policy DMD
    training corrects rollout errors, while sliding-window KV caching
    reduces accumulation of color and appearance drift. Motion dynamics
    slightly favors Base (24.0\% vs.\ 16.0\%), and text controllability
    and audio-video synchronization are near-equivalent (82--86\% Same).
\end{itemize}
Overall, the distilled online model retains the core capabilities of Base
LPM with a predictable quality redistribution: some loss in subtle
non-speech expressiveness (Listen), but meaningful gains in motion
smoothness (Speak) and long-horizon identity stability (Conversation).

%% file: 5_eval_sections/4_ablation_study.tex
\subsection{Ablation Study}
\label{sec:ablation_study}

We further study the role of different reference signals for identity preservation in LPM. To ensure a controlled comparison, all other inputs are held fixed, and only the reference images are varied. Specifically, we consider two types of references, emotion references and view references, and analyze their distinct contributions to identity fidelity.

\begin{figure}[ht]
    \centering
    \includegraphics[width=1\textwidth]{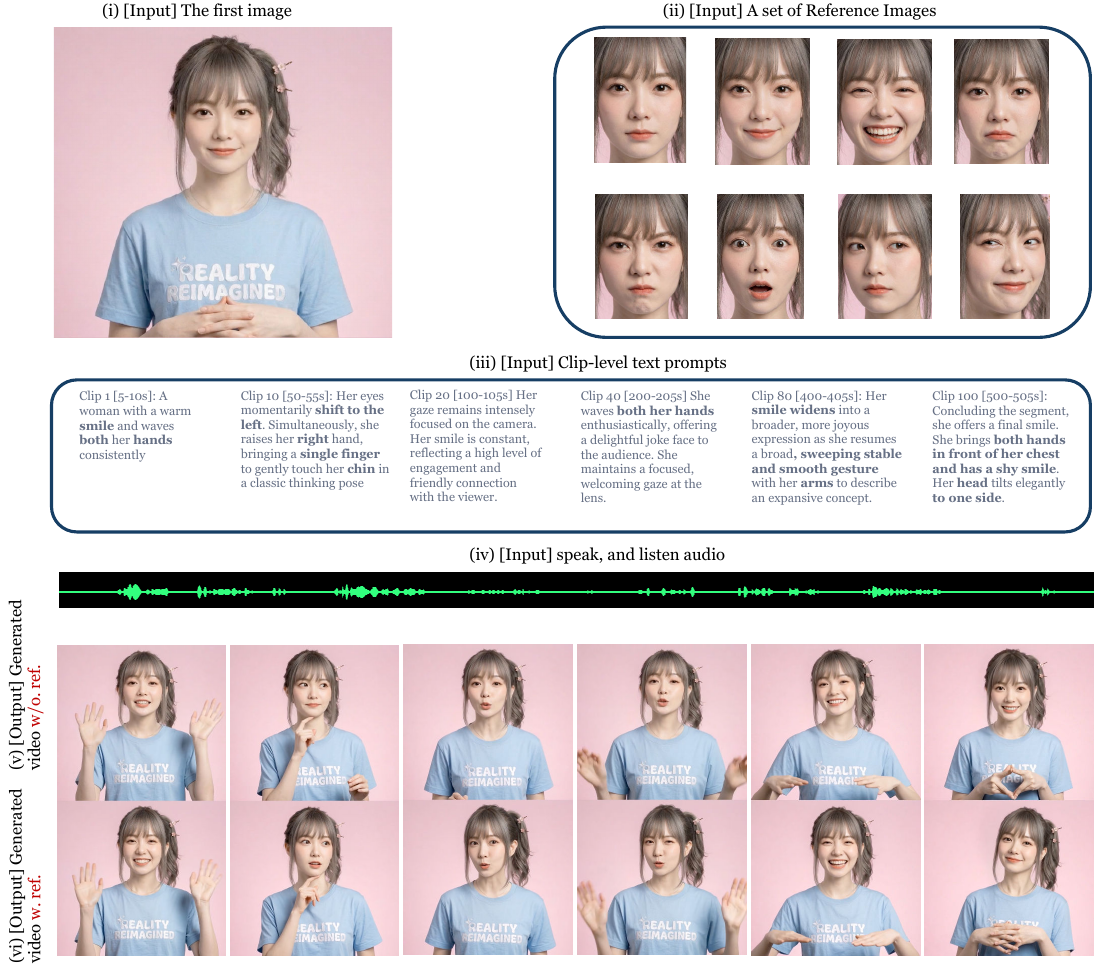}
    \caption{Ablation on emotion references of Base LPM. With all other inputs fixed, adding emotion reference images improves the preservation of fine-grained identity-related expression cues, including smiling style, dental appearance, and subtle facial micro-expressions. In contrast, without emotion references, these person-specific expressive details are less faithfully preserved.}
    \label{fig:results1}
\end{figure}

Emotion references primarily preserve fine-grained expressive traits. As shown in Figure~\ref{fig:results1}, adding emotion reference images helps the model retain identity-specific expression cues, including smiling style, dental appearance, and subtle facial micro-expressions. Without these references, the generated videos still preserve the subject's overall appearance to a large extent, but the expressions become noticeably more generic and less characteristic of the target individual. This result indicates that emotion references are particularly beneficial for maintaining person-specific facial dynamics that cannot be reliably inferred from coarse identity cues alone.

\begin{figure}[ht]
    \centering
    \includegraphics[width=1\textwidth]{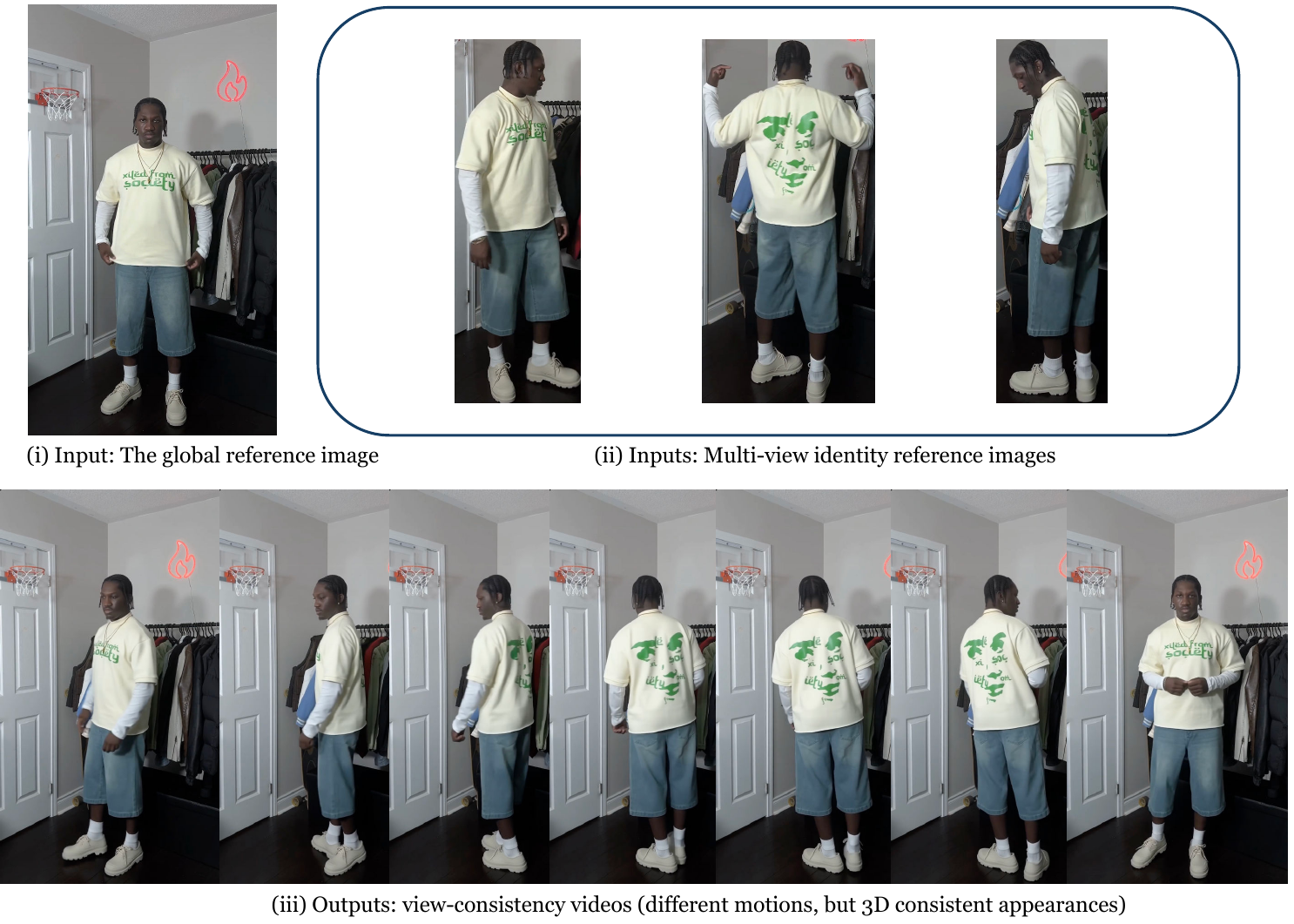}
    \caption{Ablation study on view references of Base LPM. Given a \textbf{(i)} global reference image and \textbf{(ii)} multi-view identity reference images, the model generates videos in which the subject undergoes substantial orientation changes while preserving 3D-consistent identity and appearance. This effect is evident not only in coherent body and facial structure, but also in orientation-dependent appearance details, such as the clothing logo on the back, which remains consistent as the subject turns.}
    \label{fig:viewref}
\end{figure}

In contrast, body-view references primarily improve identity consistency under subject orientation changes. Figure~\ref{fig:viewref} shows that incorporating identity references with diverse subject orientations yields substantially stronger coherence as the subject turns. The improvement is visible at multiple levels. Globally, the generated subject maintains a more stable body and facial structure throughout the motion. Locally, orientation-dependent appearance details, such as the logo on the back of the shirt, remain consistent as the subject rotates rather than being distorted or omitted. Without view references, these pose-sensitive identity cues are considerably harder to preserve, leading to weaker structural consistency and less faithful rendering of newly exposed appearance details.

Taken together, the two reference types play complementary roles. Emotion references sharpen identity-related expressive fidelity, whereas view references reinforce structural and appearance consistency under pose changes. Their combination, therefore, supports a more complete and robust form of identity preservation, which is particularly important for long-form human video generation involving both rich facial expressions and substantial body motion.

%% file: 7_discussion.tex
\section{Summary and Discussion}
\label{sec:discussion}

\paragraph{Summary.}
\emph{LPM~1.0} begins from a simple insight: human conversation is not merely the exchange of words, but a form of performance. What makes an interaction feel natural depends not only on semantic content, but on how attention, timing, reaction, and affect are continuously expressed through voice, face, and body. A conversational character, therefore, should not be evaluated solely by lip synchronization or frame realism, but by whether it appears to participate in the interaction as a socially legible actor: listening while silent, anticipating turn transitions, reacting contingently, and remaining behaviorally coherent over time \cite{cassell2001embodied}. From this perspective, the objective is not a better talking head, but a \emph{performance model}: a system that sustains the audiovisual behavior of a conversational actor over time.
In this work, our system suggests that the performance trilemma — expressive quality, real-time inference, and long-horizon stability — admits a workable resolution through systems-level co-design. In particular, conversational performance at this stage is more naturally addressed as a joint problem of data, multimodal conditioning, generation, streaming, and stabilization than as a question of a model architecture alone. The resulting system is not a complete solution to interactive character, but it demonstrates that high-quality full-duplex conversational performance can be made practical under deployable latency and stability constraints.

\paragraph{Limitations.}
This resolution is possible only because the current regime remains relatively simple. Interaction is still mostly limited to a single, camera-facing character and is weakly grounded in broader social, physical, and narrative structure. The system is not yet required to support long-horizon discourse memory, multi-party coordination, behavior in dynamically changing environments, or strong 3D and world consistency under arbitrary viewpoints and actions. As a result, specialized components can still achieve strong performance without making the overall system feel fragmented.

\paragraph{Future work.}
Looking ahead, we see three key axes for extending this work.
Along the \emph{temporal} axis, longer interactions will require discourse-level memory, persona persistence, and the ability to make current behavior coherent with prior events.
Along the \emph{social} axis, multi-party interaction introduces new challenges such as addressee tracking, gaze allocation, and group-level turn-taking~\cite{bohus2010facilitating}.
Along the \emph{physical} axis, characters situated in environments must ground their behavior in scene geometry, objects, and contact.
As these dimensions converge, the current pipeline decomposition---language generation, speech synthesis, audiovisual rendering, and online stabilization---may give way to more unified actor models that jointly determine what is said, how it is expressed, and how behavior unfolds over time.
\emph{LPM~1.0} is best viewed as a first systems-level answer to this larger problem, demonstrating that video generation can serve not only as a rendering mechanism, but as the layer through which an interactive character becomes perceptible as a participant.

%% file: 8_safety_sections/1_safety_security_responsibility.tex
\section{Safety, Security, and Responsibility}
\label{sec:safety_security_responsibility}

\paragraph{Identity Protection and Consent.}
\emph{LPM~1.0} generates human-like character video conditioned on reference images, which inherently raises concerns about unauthorized identity replication, non-consensual likeness usage, and potential violations of personality rights across jurisdictions. We take a proactive position on identity safety through both technical and procedural measures. First, \textbf{all reference images and audio used in our demonstrations, evaluations, and public-facing results are synthetically generated} from commercially accessible generative models, ensuring that no real person's likeness is reproduced without consent. This design choice significantly reduces the risk of celebrity or private-person identity leakage in our published outputs. Second, we advocate and support the adoption of explicit consent protocols in any downstream deployment: systems built on \emph{LPM~1.0} should require verified authorization before generating video conditioned on a specific individual's appearance. Moreover, we recommend that deployments maintain auditable logs linking each generation to its authorized reference source. We recognize that as performance models become more capable, the boundary between synthetic and authentic human appearance will continue to narrow. This makes identity governance not merely a legal compliance matter, but a foundational design requirement for any responsible deployment of character generation technology.

\paragraph{Misuse Prevention and Technical Safeguards.}
The ability to generate realistic, real-time human video introduces risks of malicious misuse, including non-consensual deepfakes, fraud, disinformation, and social engineering attacks. We address these risks through multiple layers of technical safeguards. First, we integrate invisible watermarking into generated video outputs, embedding provenance metadata that enables downstream detection and attribution of AI-generated content, consistent with the C2PA (Coalition for Content Provenance and Authenticity) framework. Second, we develop and release companion AI-generated content detection models trained to distinguish \emph{LPM} outputs from authentic video, supporting platform-level content moderation. Third, we implement input-level safety filtering that screens reference images, audio inputs, and text prompts against known risk categories, including NSFW content, hate speech, and violence-inciting directives, before generation proceeds. Fourth, for production deployments, we recommend tiered access control with purpose-of-use declaration and periodic audit reviews. We acknowledge that no single safeguard is sufficient in isolation; effective misuse prevention requires a comprehensive defense strategy combining technical, institutional, and regulatory mechanisms.

\paragraph{Responsible Deployment and Social Impact.}
We believe that video-generative performance models have significant potential to benefit society from enabling accessible AI companions for elderly care, mental health support, and online education, to democratizing content creation for various users, to powering inclusive educational experiences across languages and abilities. Realizing this potential responsibly requires deliberate attention to fairness, representation, and transparency. On \textbf{fairness and bias}, our training data pipeline incorporates demographic-aware quality filtering and balanced sampling to mitigate the reproduction of societal biases in generated character appearance, behavior, and interaction style; we commit to ongoing bias auditing as the model evolves. On \textbf{transparency}, we advocate that all AI-generated character video should be clearly disclosed, and we support emerging regulatory frameworks, including the EU AI Act, China's Deep Synthesis Provisions, and NIST AI Risk Management Framework, that establish disclosure and accountability standards for generative AI systems. On \textbf{dual-use governance}, we adopt a responsible release strategy: model API access is provided under license terms that explicitly prohibit use for non-consensual impersonation, political manipulation, and fraudulent misrepresentation. We are committed to iterating on these safeguards in collaboration with the broader research community, policymakers, and civil society to ensure that advances in character performance generation serve the collective good, making conversations more natural, more inclusive, and more trustworthy for everyone.

%% file: 9_contributors.tex
\section{Acknowledgement}
\label{sec:contributors}

We thank all contributors who participated in different stages of this project, including data preparation, model development, and evaluation. We also appreciate the support from our extended collaborators and previous contributors, including interns and former team members who contributed to earlier versions of this work. Their contributions have been essential to the development of this project.

We also gratefully acknowledge the open-source community and the authors of the related works cited in this paper, whose publicly available models, tools, and datasets have been invaluable to this work.